\documentclass[letterpaper, 10 pt, conference]{ieeeconf}  

\IEEEoverridecommandlockouts                              
\usepackage{url}
\usepackage{epsfig} 
\usepackage{multirow}
\usepackage{float}

\title{\LARGE \bf
Learn Convolutional Neural Network for Face Anti-Spoofing
}

\author{Jianwei Yang, Zhen Lei and Stan Z. Li
\thanks{Jianwei Yang, Zhen Lei, and Stan Z. Li are with Center for Biometrics and Security Research \& National Laboratory of Pattern Recognition, Institute of Automation, Chinese Academy of Sciences, China. Email: \{jwyang, zlei, szli\}@cbsr.ia.ac.cn
}
}

\begin{document}

\maketitle
\thispagestyle{empty}
\pagestyle{empty}

\begin{abstract}

Though having achieved some progresses, the hand-crafted texture features, e.g., LBP \cite{DBLP:conf/IJCB/Maatta}, LBP-TOP \cite{de2013lbp} are still unable to capture the most discriminative cues between genuine and fake faces. In this paper, instead of designing feature by ourselves, we rely on the deep convolutional neural network (CNN) to learn features of high discriminative ability in a supervised manner. Combined with some data pre-processing, the face anti-spoofing performance improves drastically. In the experiments, over 70\% relative decrease of Half Total Error Rate (HTER) is achieved on two challenging datasets, CASIA \cite{Face_Anti_Spoofing_DataBase_ZhiweiZhang_2012} and REPLAY-ATTACK \cite{Face_Anti_Spoofing_DataBase_Chingovska_2012} compared with the -state-of-the-art. Meanwhile, the experimental results from inter-tests between two datasets indicates CNN can obtain features with better generalization ability. Moreover, the nets trained using combined data from two datasets have less biases between two datasets.

\end{abstract}

\section{INTRODUCTION}

\begin{figure}
\begin{minipage}[b]{0.2\linewidth}
\centering
\centerline{\epsfig{figure=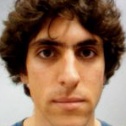,height=2.0cm, width = 2.0cm}}
\end{minipage}
\hfill
\begin{minipage}[b]{0.2\linewidth}
\centering
\centerline{\epsfig{figure=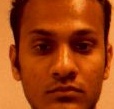,height=2.0cm, width = 2.0cm}}
\end{minipage}
\hfill
\begin{minipage}[b]{0.2\linewidth}
\centering
\centerline{\epsfig{figure=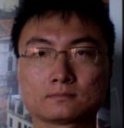,height=2.0cm, width = 2.0cm}}
\end{minipage}
\hfill
\begin{minipage}[b]{0.2\linewidth}
\centering
\centerline{\epsfig{figure=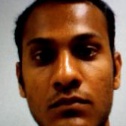,height=2.0cm, width = 2.0cm}}
\end{minipage}

\begin{minipage}[b]{0.2\linewidth}
\centering
\centerline{\epsfig{figure=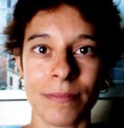,height=2.0cm, width = 2.0cm}}
\end{minipage}
\hfill
\begin{minipage}[b]{0.2\linewidth}
\centering
\centerline{\epsfig{figure=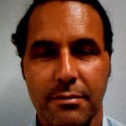,height=2.0cm, width = 2.0cm}}
\end{minipage}
\hfill
\begin{minipage}[b]{0.2\linewidth}
\centering
\centerline{\epsfig{figure=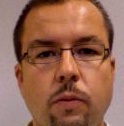,height=2.0cm, width = 2.0cm}}
\end{minipage}
\hfill
\begin{minipage}[b]{0.2\linewidth}
\centering
\centerline{\epsfig{figure=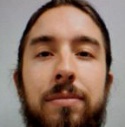,height=2.0cm, width = 2.0cm}}
\end{minipage}
\caption{Genuine and fake face images from REPLAY-ATTACK dataset.}
\label{Fig_1}
\end{figure}

Face anti-spoofing, as a security measure for face recognition system, are drawing increasing attentions in both academic and industrial fields. However, due to the diversity of spoofing types, including print-attacks, replay-attacks, mask-attacks, etc., it is still a difficult work to distinguish various fake faces. In Fig.~\ref{Fig_1}, some randomly sampled genuine and fake face images are shown to evaluate the anti-spoofing ability of our eyes. Among all the face images, three are genuine and five are fake \footnote{The second and third images in the top row, and the third image in the bottom row are genuine}. Admittedly, no obvious visual cues are available for us to pick the genuine face images from the gallery.

Recently, researchers are devoted to come up with more generalized and discriminative features for face anti-spoofing, such as LBP \cite{DBLP:conf/IJCB/Maatta}\cite{Face_Anti_Spoofing_JianweiYang_2013}, HOG \cite{komulainen2013context}\cite{Face_Anti_Spoofing_JianweiYang_2013}, LBP-TOP \cite{de2013lbp}, DoG \cite{Face_Anti_Spoofing_XiaoyangTan_ECCV_2010} \cite{DBLP:conf/icip/PeixotoMR11}, etc. In general, these features are all called hand-crafted features because they are designed manually. In this paper, however, we exploit deep convolutional neural network (CNN) for face anti-spoofing. To the best of our knowledge, this is the first attempt. Compared with above hand-crafted features, the features learned from CNN are able to catch more discriminative cues in a data-driven manner. More importantly, according to the experimental results, it has the potential to learn more general features for various spoofing types.

\section{RELATED WORKS}

Due to the diversity of spoofing attacks, existing face anti-spoofing approaches can be mainly categorized into four groups: texture based, motion based, 3D-shape based and multi-spectral reflectance based. Besides, some other works combined two or more of these methods to improve the anti-spoofing performance.

\subsubsection{Texture-based Anti-Spoofing}

In \cite{Face_Anti_Spoofing_JianweiLi_Fourier_2004}, Li et al. proposed a method based on the analysis of Fourier spectra. In this method, it is assumed that the photographs contain fewer high frequency components compared with genuine faces. In \cite{Face_Anti_Spoofing_XiaoyangTan_ECCV_2010}, Tan et al. used a variational retinex-based method and the difference-of-Gaussian (DoG) filers to extract latent reflectance features on face images. Inspired by Tan's work, Peixoto et al. \cite{DBLP:conf/icip/PeixotoMR11} combined the DoG filters and standard Sparse Logistic Regression Model for anti-spoofing under extreme illuminations. After that, M\"{a}\"{a}tt\"{a} et al. \cite{DBLP:conf/IJCB/Maatta} proposed to use LBP features for anti-spoofing, which outperformed previous methods on the NUAA Photograph Imposter Database \cite{DBLP:conf/eccv/TanLLJ10}. Furthermore, its efficiency on the REPLAY-ATTACK database was presented in \cite{Face_Anti_Spoofing_DataBase_Chingovska_2012}.

In \cite{de2013lbp}, Pereira et al. used a spatio-temporal texture feature called Local Binary Patterns from Three Orthogonal Planes (LBP-TOP). According to the experimental results on the REPLAY-ATTACK database, it outperformed the LBP-based method in \cite{DBLP:conf/IJCB/Maatta}. In \cite{Face_Anti_Spoofing_Pereira_2013}, it is shown that LBP and LBP-TOP features are applicable in intra-database protocol. However, the countermeasures performance degraded much in a more realistic scenario, i.e., inter-database protocol. The reason for the low generalization ability of texture features was partially explained in the paper \cite{Face_Anti_Spoofing_JianweiYang_2013}. The authors found many factors may affect the textures on a face image, including abnormal shadings, highlights, device noises, etc. Actually, the usage of texture features are not confined in above papers. In the $1^{st}$ and $2^{nd}$ competition on 2D face anti-spoofing \cite{Face_Anti_Spoofing_Competition_2011} \cite{Face_Anti_Spoofing_Competition_2013}, most of the teams used textures as clues for anti-spoofing.

\subsubsection{Motion-based Anti-Spoofing}

Beyond the texture features, motion is another cues for face anti-spoofing. In \cite{DBLP:conf/icb/SunPWL07}\cite{DBLP:conf/iccv/PanSWL07}, Pan et al. used eye blinking for face anti-spoofing. In their method, a conditional random field was constructed to model different stages of eye blinking. In \cite{DBLP:journals/tifs/KollreiderFFB07}, Kollreider et al. used lip movement classification and lip-reading for the purpose of liveness detection. The system requires users to utter a sequence of words, and then verify whether the observed lip dynamics fit in the words. Furthermore, Chetty et al. \cite{DBLP:conf/fuzzIEEE/Chetty10}\cite{Chetty05} proposed a multi-modal approach to aggrandize the difficulty of spoofing attacks. It determined the liveness by verifying the fitness between video and audio signals.

On the other hand, some previous works focused on physical motions for anti-spoofing. In \cite{DBLP:conf/IASP/W.Bao}, Bao et~al. presented a method using optical flow fields to distinguish 2-D planar photography attacks and 3-D real faces. Similarly, Kollreider et~al. \cite{DBLP:conf/autoid/KollreiderFB05} \cite{Face_Anti_Spoofing_MotionAnalysis_Kollreider_2007} also relied their method on optical flow analysis. The method is based on the assumption that a 3-D face generates a special 2-D motion which is higher at central face parts (e.g. nose) compared to the outer face regions (e.g. ears). More recently, Anjos et~al. proposed to recognize spoofing attacks based on the correlation between optical flows in foreground and background regions \cite{anjos2013motion}. At the same time, Yang et~al. presented a counter measure to replay attacks based on the correlations among optical magnitude/phase sequences from 11 regions, which won the first place after combining with a texture-based method \cite{Face_Anti_Spoofing_Competition_2013}. Besides, Kollreider et~al. \cite{DBLP:conf/Biometrics/Kollreider} used eye-blinking and face movements for detecting liveness in an interaction scenario.

\subsubsection{3D Shape-based Anti-Spoofing}

In \cite{de2012moving}, Marsico et~al. proposed a method for moving face anti-spoofing based on 3D projective invariants. However, this method can merely cope with photo attacks without warping, because the coplanar assumption is invalid for warped photos. Though warped photos do not satisfy the coplanar constrains as real face, there are still some intrinsic differences between them. In \cite{wang2013face}, the authors proposed to recover sparse 3D shapes for face images to detect various photo attacks. The performance on different warping types (none, vertically and horizontally) are evaluated, which showed that the method worked perfectly under both intra-database protocols and inter-database protocols. However, both methods will fail when coping with 3D mask spoofing, such as the 3D Mask Attack database (3DMAD) collected by Erdogmus et~al. \cite{erdogmus2013spoofing}.

\subsubsection{Multi-Spectral Reflectance-based Anti-Spoofing}

The multi-spectral methods utilize the illuminations beyond visual spectrum for detect spoofing attacks. In \cite{DBLP:conf/CVBVS/Pavlidis} and \cite{DBLP:conf/fgr/ZhangYLL11}, the authors selected proper working spectrums so that the reflectance differences between genuine and fake faces increased. Different from the methods directly using reflection intensities, a gradient-based multi-spectral method for face anti-spoofing was proposed in \cite{hou2013multispectral}. The authors studied three illumination-robust features and evaluated the performance on different spectral bands. However, these methods need extra devices to capture face images under the invisible lights, thus it is unpractical to deploy such devices to the most of recent FR systems, which are merely based on RGB color face images.

Moreover, some works combined two or more of above four kinds of approaches. In \cite{Face_Anti_Spoofing_Pan_Eyeblink_2007}, Pan et~al. integrated scene context into their earlier eye blinking based anti-spoofing scheme. However, the so-called scene context is non-existed in many cases, such as the PRINT-ATTACK database. Toward the PRINT-ATTACK database, Tronci et~al. employed motion, texture and liveness \cite{tronci2011fusion} and achieved perfect performance on development set and test set. On the same database, Yan et~al. \cite{yan2012face} explored multiple scenic clues, including non-rigid motion, face-background consistency and image banding effect, to detect the spoofing attacks, which achieved 100\% accuracy on the test set. Recently, Chingovska et~al proposed to integrate face recognition module into anti-spoofing system in score-level and feature level \cite{anjos2013anti}.

\section{METHOD}

\subsection{Data Preparation}

\subsubsection{face localization}
\label{face_localization}

\begin{figure}[!hb]
\begin{minipage}[b]{0.18\linewidth}
\centering
\centerline{\epsfig{figure=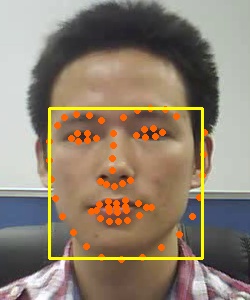,height=2.16cm, width = 1.8cm}}
\end{minipage}
\hfill
\begin{minipage}[b]{0.18\linewidth}
\centering
\centerline{\epsfig{figure=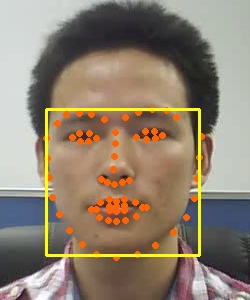,height=2.16cm, width = 1.8cm}}
\end{minipage}
\hfill
\begin{minipage}[b]{0.18\linewidth}
\centering
\centerline{\epsfig{figure=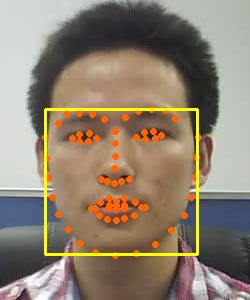,height=2.16cm, width = 1.8cm}}
\end{minipage}
\hfill
\begin{minipage}[b]{0.18\linewidth}
\centering
\centerline{\epsfig{figure=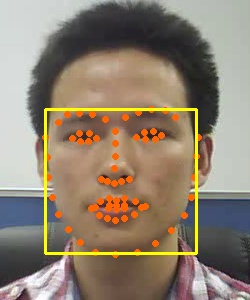,height=2.16cm, width = 1.8cm}}
\end{minipage}
\hfill
\begin{minipage}[b]{0.18\linewidth}
\centering
\centerline{\epsfig{figure=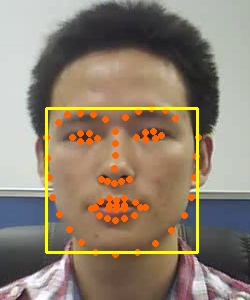,height=2.16cm, width = 1.8cm}}
\end{minipage}
\caption{Face alignment process on an example image. From left to right, the face location is gradually refined based on face landmarks.}
\label{Fig_3}
\end{figure}

Before face anti-spoofing, face localization is indispensable. In previous works, a common face detector, e.g., Viola-Jones in OpenCV, is enough for this task. However, such rectangle-wise detector cannot provide precise face locations. Therefore, we implement the face alignment algorithm proposed in \cite{renface} after a common Viola-Jones face detection. In the training stage, we extracts a set of local binary features, which are then used to learn a linear regressors in each cascade. During testing, an initial rectangle is provided by face detector, followed by a cascaded regression for the final output, i.e. a group of face landmarks. After obtaining the landmarks, their bounding box is regarded as the final face location. As shown in Fig.~\ref{Fig_3}, the initial rectangle is refined gradually based on face landmarks to obtain a more precise face location.

\subsubsection{spatial augmentation}

Different from some other face-oriented algorithms, such as face detection and face recognition, face anti-spoofing is more like an image quality diagnosing issue. Beyond the conventional face region, the backgrounds are helpful for the classification as well. In \cite{Face_Anti_Spoofing_Competition_2011}, the team UOULU exploited background region for feature extraction, and achieved best performance in the competition. In \cite{Face_Anti_Spoofing_JianweiYang_2013}, the authors enlarged the conventional face region to contain a part of background, and proved the positive role of background region with various feature types. Inspired by their works, we also propose to enlarge the face region to contain background region. However, the difference is that we tend to use more backgrounds in our method. Though it is shown in \cite{Face_Anti_Spoofing_JianweiYang_2013} that extra background made no difference on the face anti-spoofing performance, we argue that the hand-crafted features the author used encounter bottlenecks to exploit more information from background regions. Alternatively, the CNN is more capable of learning discriminative features from backgrounds.

\begin{figure}
\begin{minipage}[b]{0.18\linewidth}
\centering
\centerline{\epsfig{figure=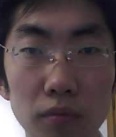,height=1.6cm, width = 1.6cm}}
\end{minipage}
\hfill
\begin{minipage}[b]{0.18\linewidth}
\centering
\centerline{\epsfig{figure=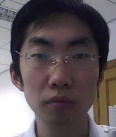,height=1.6cm, width = 1.6cm}}
\end{minipage}
\hfill
\begin{minipage}[b]{0.18\linewidth}
\centering
\centerline{\epsfig{figure=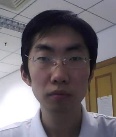,height=1.6cm, width = 1.6cm}}
\end{minipage}
\hfill
\begin{minipage}[b]{0.18\linewidth}
\centering
\centerline{\epsfig{figure=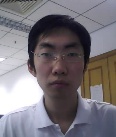,height=1.6cm, width = 1.6cm}}
\end{minipage}
\hfill
\begin{minipage}[b]{0.18\linewidth}
\centering
\centerline{\epsfig{figure=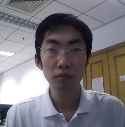,height=1.6cm, width = 1.6cm}}
\end{minipage}

\begin{minipage}[b]{0.18\linewidth}
\centering
\centerline{\epsfig{figure=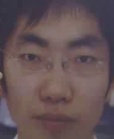,height=1.6cm, width = 1.6cm}}
\end{minipage}
\hfill
\begin{minipage}[b]{0.18\linewidth}
\centering
\centerline{\epsfig{figure=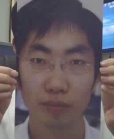,height=1.6cm, width = 1.6cm}}
\end{minipage}
\hfill
\begin{minipage}[b]{0.18\linewidth}
\centering
\centerline{\epsfig{figure=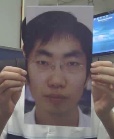,height=1.6cm, width = 1.6cm}}
\end{minipage}
\hfill
\begin{minipage}[b]{0.18\linewidth}
\centering
\centerline{\epsfig{figure=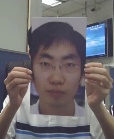,height=1.6cm, width = 1.6cm}}
\end{minipage}
\hfill
\begin{minipage}[b]{0.18\linewidth}
\centering
\centerline{\epsfig{figure=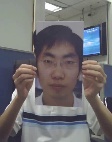,height=1.6cm, width = 1.6cm}}
\end{minipage}

\begin{minipage}[b]{0.18\linewidth}
\centering
\centerline{\epsfig{figure=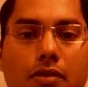,height=1.6cm, width = 1.6cm}}
\end{minipage}
\hfill
\begin{minipage}[b]{0.18\linewidth}
\centering
\centerline{\epsfig{figure=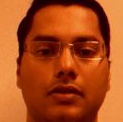,height=1.6cm, width = 1.6cm}}
\end{minipage}
\hfill
\begin{minipage}[b]{0.18\linewidth}
\centering
\centerline{\epsfig{figure=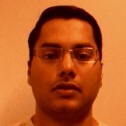,height=1.6cm, width = 1.6cm}}
\end{minipage}
\hfill
\begin{minipage}[b]{0.18\linewidth}
\centering
\centerline{\epsfig{figure=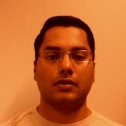,height=1.6cm, width = 1.6cm}}
\end{minipage}
\hfill
\begin{minipage}[b]{0.18\linewidth}
\centering
\centerline{\epsfig{figure=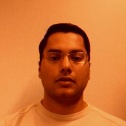,height=1.6cm, width = 1.6cm}}
\end{minipage}

\begin{minipage}[b]{0.18\linewidth}
\centering
\centerline{\epsfig{figure=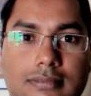,height=1.6cm, width = 1.6cm}}
\end{minipage}
\hfill
\begin{minipage}[b]{0.18\linewidth}
\centering
\centerline{\epsfig{figure=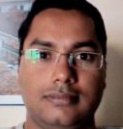,height=1.6cm, width = 1.6cm}}
\end{minipage}
\hfill
\begin{minipage}[b]{0.18\linewidth}
\centering
\centerline{\epsfig{figure=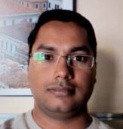,height=1.6cm, width = 1.6cm}}
\end{minipage}
\hfill
\begin{minipage}[b]{0.18\linewidth}
\centering
\centerline{\epsfig{figure=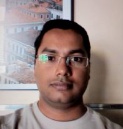,height=1.6cm, width = 1.6cm}}
\end{minipage}
\hfill
\begin{minipage}[b]{0.18\linewidth}
\centering
\centerline{\epsfig{figure=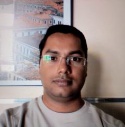,height=1.6cm, width = 1.6cm}}
\end{minipage}
\caption{Sample images with different spatial scales are shown. The first two rows are from CASIA dataset and the two rows at bottom are from REPLAY-ATTACK dataset. The even rows are genuine, and odd rows are fake.}
\label{Fig_2}
\end{figure}

As shown in Fig.~\ref{Fig_2}, we prepare the input images with five scales. Images corresponding to the first scale merely contain face region. With the increase of scale, images contain more background regions. As for CASIA-FASD dataset, we can easily find that fake images in large-scale contain boundaries of photographs compared with genuine images, which should be exploited as discriminative cues for anti-spoofing. In another case as REPLAY-ATTACK dataset, though fake images have no boundary cues, they contains blurred edges and probable abnormal specular reflections caused by re-capturing compared with genuine samples in whole images \cite{Face_Anti_Spoofing_JianweiYang_2013}.

\subsubsection{temporal augmentation}

Besides spatial augmentations above, we also propose to augment the data temporally. Multiple frames are expected to improve the anti-spoofing performance due to more informative data. This has been proved by \cite{de2013lbp} to some extent, in which a spatial-temporal texture feature was extracted from consecutive frames. When fed more than one frame, the CNN can not only learn the spatial features, but also temporal features for anti-spoofing. In this paper, we train CNN model using both single frame and multiple frames, and figure out whether multiple frames are helpful for CNN to learn more discriminative features. 

\subsection{Feature Learning}

\begin{figure}
\begin{minipage}[b]{1\linewidth}
\centering
\centerline{\epsfig{figure=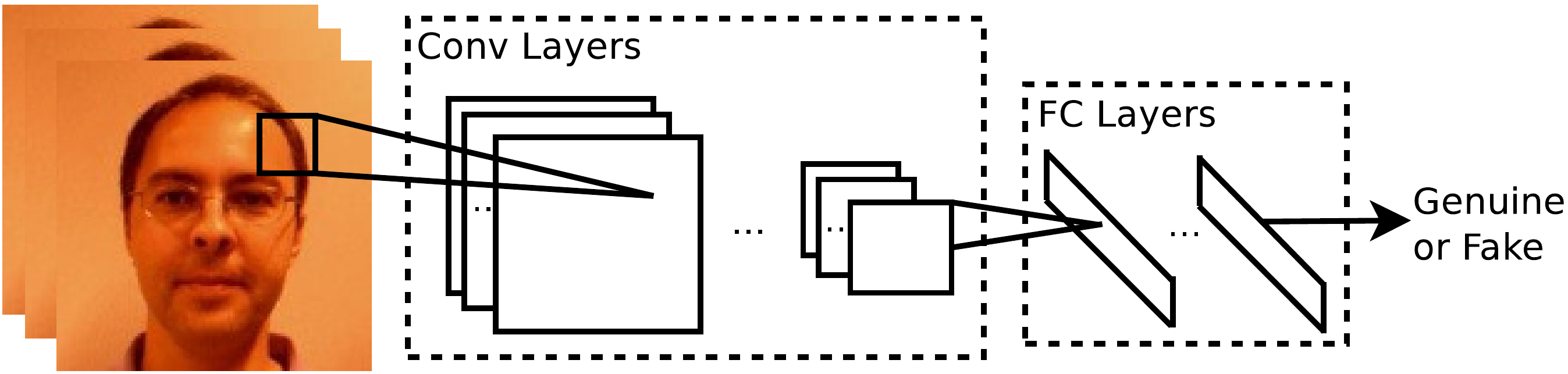,height=2.1cm}}
\end{minipage}
\caption{Brief illustration of CNN architecture used in this paper.}
\label{Fig_4}
\end{figure}

In this paper, we implement a canonical CNN structure for learning features. Specifically, we adopt the configuration in \cite{DBLP:conf/nips/KrizhevskySH12}, which won the ImageNet large scale visual recognition challenge (LSVRC) in 2012. A brief illustration of the CNN structure is presented in Fig.~\ref{Fig_4}. In the network, there are five convolutional (Conv) layers, followed by three fully-connected (FC) layers. In the convolutional layers, response-normalization layers are used for the outputs of the first and second convolutional layers. The max-pooling layers are plug to process the outputs of the first, second and the last convolutional layers. The ReLU non-linearity is applied to the output of every convolutional and fully-connected layer. To avoid over-fitting, the first two fully-connected layers are followed by two dropout layers, and the last layer, i.e. output layer is followed by softmax.

\subsection{Classification}

After learning the CNNs, we extract the features from the last fully-connected layer. Then, support vector machine (SVM) is used to learn the classifiers from train data for face anti-spoofing. In this paper, the LibSVM toolkit \cite{Chang01libsvm:a} is used.

\section{EXPERIMENTS}

\subsection{Settings}

In our experiments, to organize the data, we first detect faces in images via detector from OpenCV. After that, we use the method in Sec.~\ref{face_localization} to refine the face regions as the bounding boxes of face landmarks. Then, the bounding boxes are reset to contain most of the face regions, as shown in the first column Fig.~\ref{Fig_2}. To make use of the information in backgrounds, we enlarge the original ones with re-scaling ratios \{1.4, 1.8, 2.2, 2.6\}. Finally, all input images are resized to 128 $\times$ 128. Besides above spatial augmentations, we use consecutive frames to augment the data temporally. In our experiments, the number of input face images is from one to three. For the CNN, we use Caffe toolbox \cite{Jia13caffe} and adopt a commonly used structure, which was ever used in \cite{DBLP:conf/nips/KrizhevskySH12}. In the training of CNN, the learning rate is 0.001; decay rate is 0.001; and the momentum during training is 0.9. Before fed into the CNN, the data are first centralized by the mean of training data. These parameters are constant in our experiments. Given the learned feature, we use SVM with RBF kernel to train classifiers for antis-spoofing.

\subsection{Datasets}

In this paper, the experiments are implemented on two datasets, CASIA and REPLAY-ATTACK datasets. In these two datasets, various spoofing types are simulated. Followings are the brief introduction of two databases:
\begin{itemize}
  \item CASIA database \cite{Face_Anti_Spoofing_DataBase_ZhiweiZhang_2012}: This database contains 50 subjects in total. For each subject, the genuine faces are captured under three qualities. The spoofing images are fabricated by implementing three kind of attacks, i.e., warped photo attack, cut photo attack and electronic screen attack in three qualities, respectively. As a result, each subject has 12 sequences (3 genuine and 9 fake ones). The overall number of sequences in the database is 600.
  \item REPLAY-ATTACK database \cite{Face_Anti_Spoofing_DataBase_Chingovska_2012}: It also contains 50 subjects. For each subject, four genuine video sequences are collected in front of two backgrounds. Similar to CASIA, three spoofing types are used, including print attack, digital photo attack, and video attack. The spoofing sequences are captured under fixed and hand-hold conditions. The overall number of sequences in the database is 1200.
\end{itemize}

\subsection{Protocol}

To make fair comparison with recent works, we use the Half Total Error Rate (HTER) to report the performance. After training, the development set is used to determine the threshold corresponding to Equal Error Rate (EER). Then the threshold is used for the computation of HTER on test set. Similar to \cite{Face_Anti_Spoofing_Pereira_2013}, we divide the training set in CASIA dataset into five folds, and one of them is used as development set and the others for training. The final performance is obtained by averaging the results from five cross validations. In REPLAY-ATTACK dataset, the development set is already given. There is no need to divide the original sets.

Along with the protocols in \cite{Face_Anti_Spoofing_Pereira_2013}, we conduct intra-test on each dataset, and inter-test to evaluate the generalization ability of our method. Moreover, we also combine two datasets to evaluate our method.

\subsection{Results of Intra-test}

\subsubsection{Test on CASIA dataset}

\begin{table*}[!htb]
\caption{Intra-test Results on CASIA dataset. The EERs and HTERs are presented for different data augmentations.}
\label{TB_CASIA_INTRA_OURS}
\center
\begin{tabular}{|c|c|c|c|c|c|c|c|c|c|c|c||c|c|}
\cline{3-14}
\multicolumn{2}{c|}{}       & \multicolumn{12}{c|}{Scale}     \\
\cline{3-14}
\multicolumn{2}{c|}{} & \multicolumn{2}{c|}{1} & \multicolumn{2}{c|}{2} & \multicolumn{2}{c|}{3} & \multicolumn{2}{c|}{4} & \multicolumn{2}{c|}{5} &  \multicolumn{2}{c|}{Mean}\\
\cline{3-14}
\multicolumn{2}{c|}{} & dev & test & dev & test & dev & test & dev & test & dev & test & dev & test \\
\hline
\multirow{3}{*}{Frame} & 1 & 7.06 & 7.38 & 5.44 & 5.58 & 4.92 & 5.09 & 5.81 & 5.84 & 6.98 & 6.99 & 6.04 & 6.18 \\ 

                       & 2 & 7.50 & 8.05 & 4.87 & 4.95 & 5.51 & 4.95 & 6.41 & 5.53 & 8.12 & 7.53 & 6.48 & 6.20 \\ 

                       & 3 & 7.80 & 7.46 & 5.61 & 5.69 & 4.64 & 5.21 & 4.68 & 5.60 & 8.09 & 7.95 & 6.16 & 6.38 \\ 
\hline
\hline
& Mean & 7.45 & 7.63 & 5.31 & 5.41 & 5.02 & 5.08 & 5.63 & 5.66 & 7.73 & 7.49 & 6.23 & 6.25\\
\hline
\end{tabular}
\end{table*}

\begin{table*}[!htb]
\caption{Intra-test Results on REPLAY-ATTACK dataset.}
\label{TB_REPLAY_ATTACK_INTRA_OURS}
\centering
\begin{tabular}{|c|c|c|c|c|c|c|c|c|c|c|c||c|c|}
\cline{3-14}
\multicolumn{2}{c|}{}       & \multicolumn{12}{c|}{Scale}     \\
\cline{3-14}
\multicolumn{2}{c|}{} & \multicolumn{2}{c|}{1} & \multicolumn{2}{c|}{2} & \multicolumn{2}{c|}{3} & \multicolumn{2}{c|}{4} & \multicolumn{2}{c|}{5} &  \multicolumn{2}{c|}{Mean}\\
\cline{3-14}
\multicolumn{2}{c|}{} & dev & test & dev & test & dev & test & dev & test & dev & test & dev & test \\
\hline
\multirow{3}{*}{Frame} & 1 & 6.10 & 2.14 & 3.51 & 3.20 & 4.47 & 3.13 & 3.41 & 2.29 & 4.14 & 2.53 & 4.33 & 2.66 \\ 

                       & 2 & 8.72 & 2.57 & 2.54 & 3.09 & 3.74 & 2.81 & 3.55 & 2.98 & 3.77 & 2.55 & 4.46 & 2.80 \\ 

                       & 3 & 7.17 & 2.19 & 3.50 & 3.28 & 3.71 & 2.86 & 3.05 & 2.32 & 3.45 & 2.21 & 4.18 & 2.57 \\ 
\hline
\hline
& Mean & 7.33 & 2.30 & 3.18 & 3.19 & 3.97 & 2.93 & 3.37 & 2.53 & 3.82 & 2.43 & 4.32 & 2.68 \\
\hline
\end{tabular}
\end{table*}

We test our method on CASIA dataset in five different spatial scales from one frame to three frames. In Table~\ref{TB_CASIA_INTRA_OURS}, the HTERs on development set and test set are shown. In the table, the average performance over scales and frames are presented al well. As we can see, with the increase of spatial scale, the anti-spoofing model perform better than that of original scale, and achieves the best when the scale is equal to 3 averagely. These results indicate the positive effect of background region on face anti-spoofing task. Actually, similar claim has been proved in \cite{Face_Anti_Spoofing_JianweiYang_2013}. However, the difference is that images corresponding to the best scale in this paper are larger than that in \cite{Face_Anti_Spoofing_JianweiYang_2013}, which shows the CNN can extract more useful information from the background region compared with the hand-crafted features. However, when the scale reaches 5, the performance degrades slightly. One possible reason is that the diversity of background region weakens the positive effect. As for the number of frames used, the model trained using one frame outperform gently the models trained with more than one frames in average. However, when reviewing the results closely, we find the best performance is obtained by using two frames with scale 2. This specific result indicates multi-frame input is positive in certain cases. 

\begin{figure*}
\begin{minipage}[b]{0.3\linewidth}
\centering
\centerline{\epsfig{figure=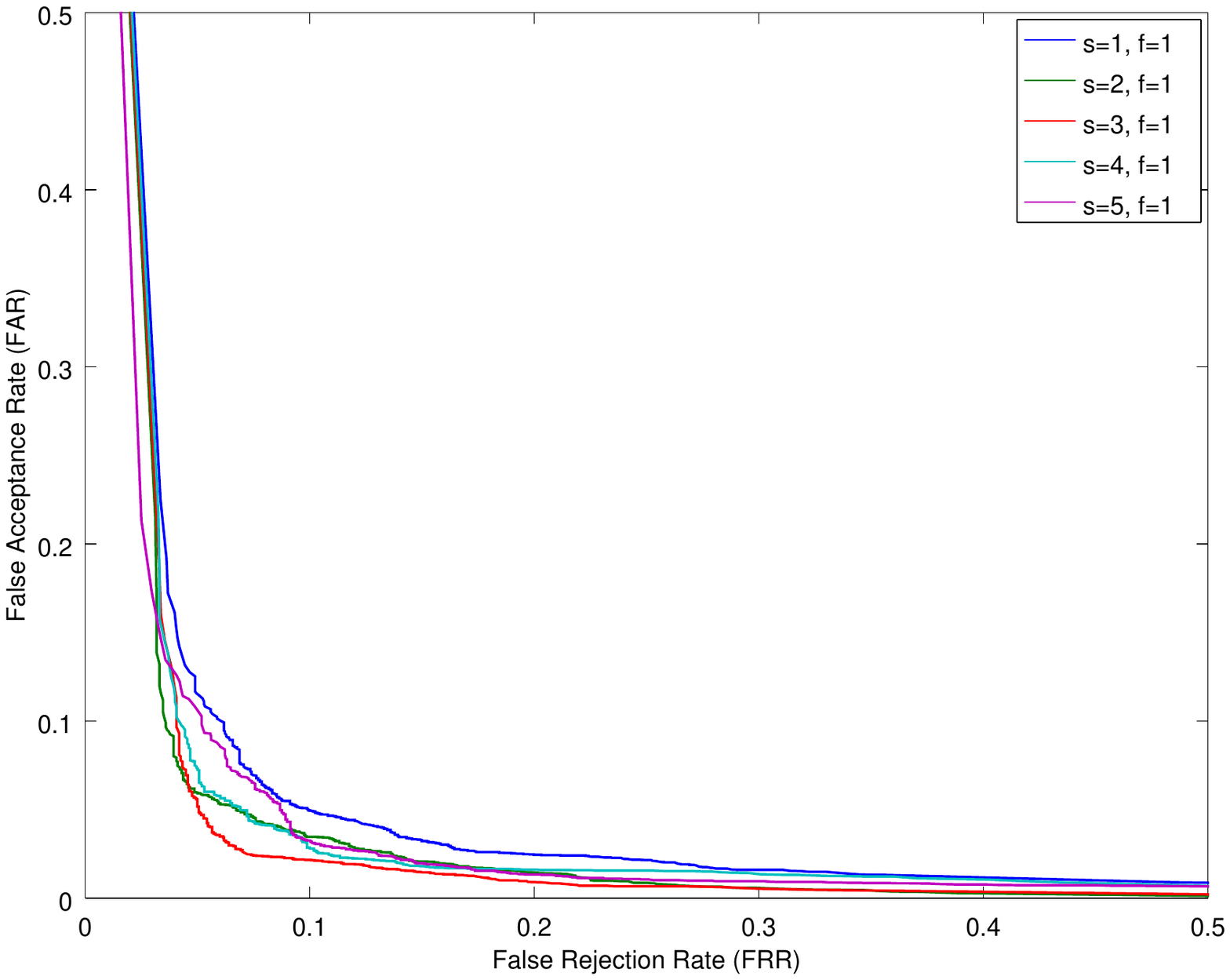,height=3.9cm, trim = 0 3in 0 3in 0}}
\end{minipage}
\hfill
\begin{minipage}[b]{0.3\linewidth}
\centering
\centerline{\epsfig{figure=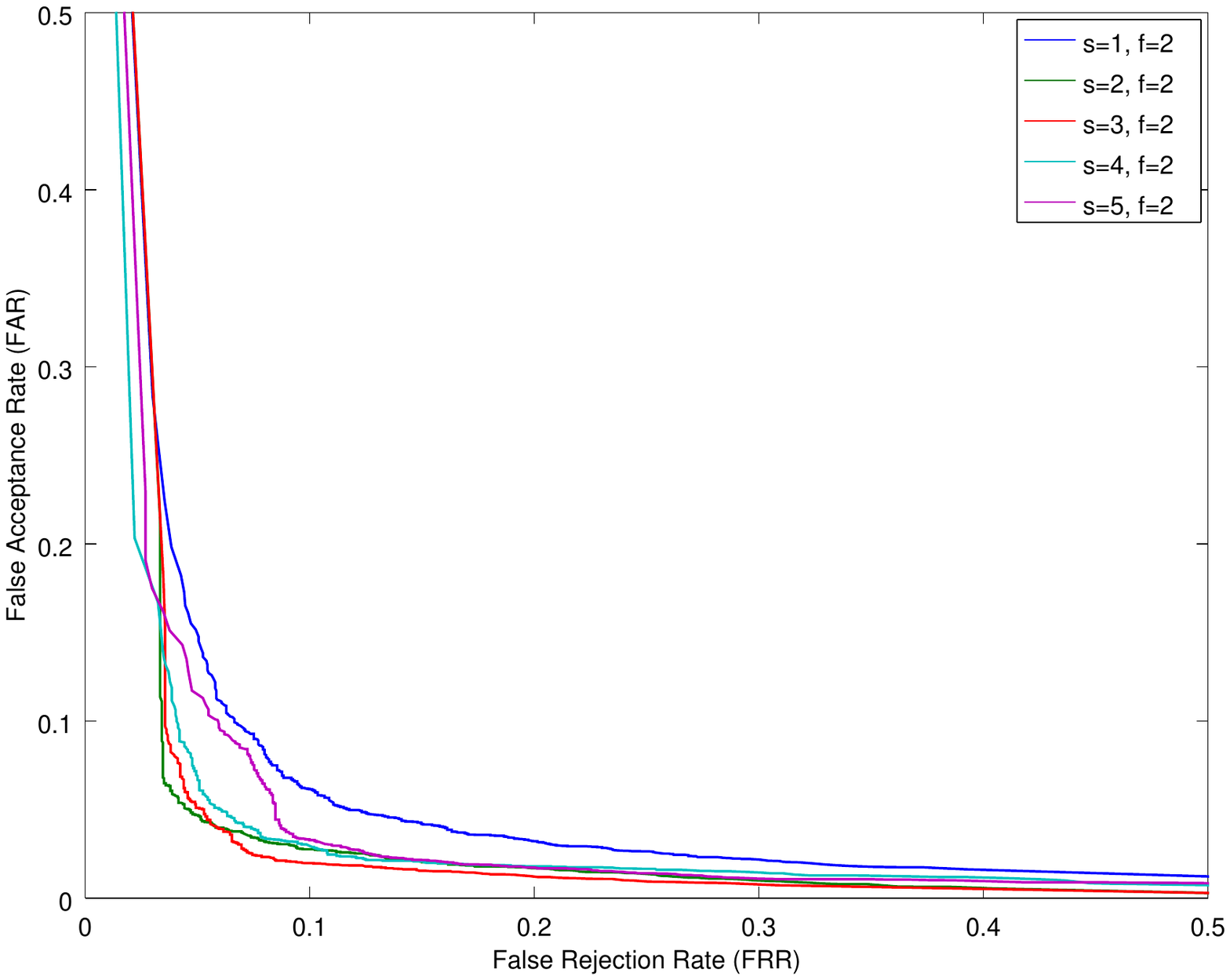,height=3.9cm, trim = 0 3in 0 3in 0}}
\end{minipage}
\hfill
\begin{minipage}[b]{0.3\linewidth}
\centering
\centerline{\epsfig{figure=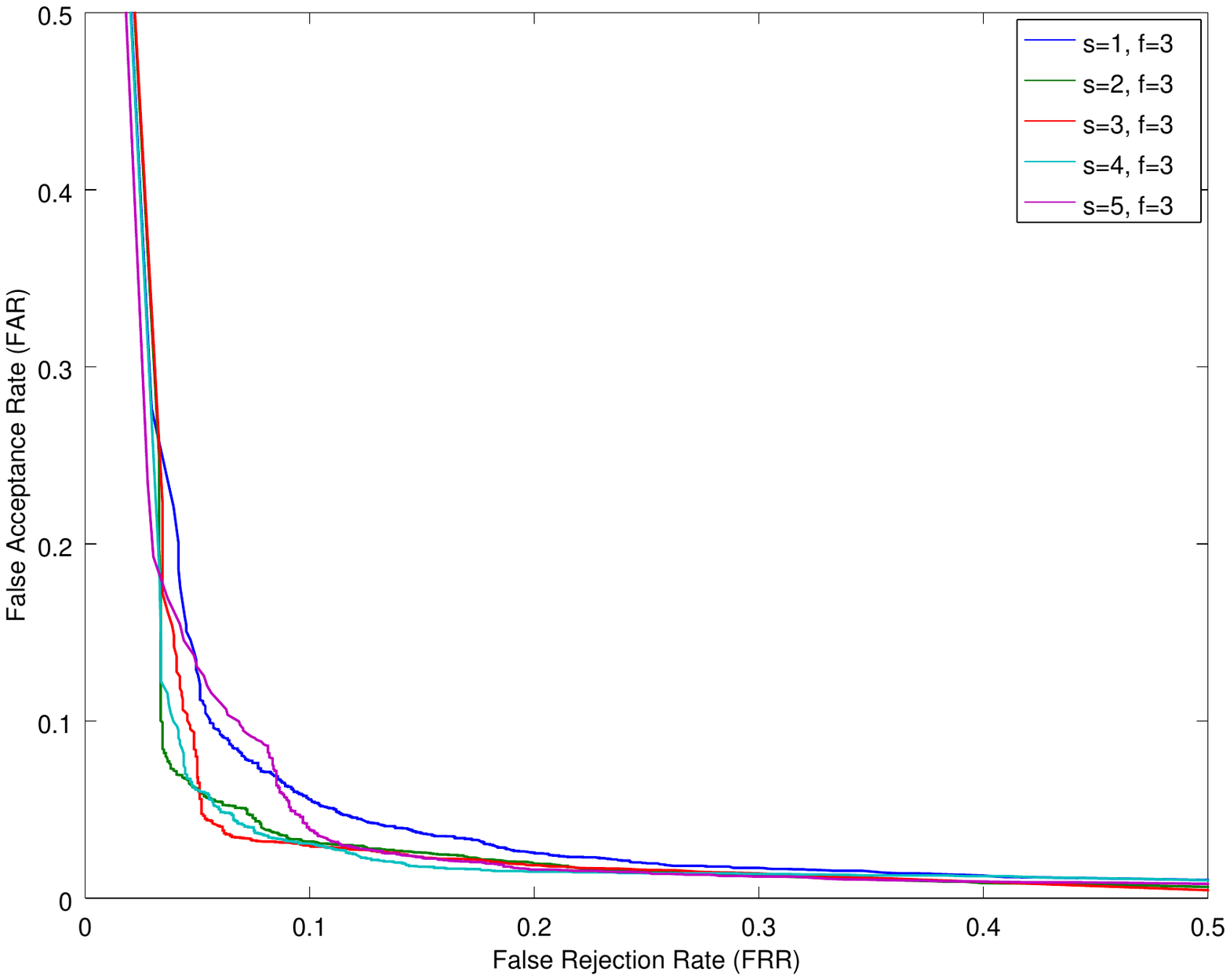,height=3.9cm, trim = 0 3in 0 3in 0}}
\end{minipage}
\caption{ROC curves for different data augmentations on CASIA dataset. From left to right, the curves are obtained from models trained using one frame, two frames and three frames, respectively.}
\label{Fig_5}
\end{figure*}

For details, we show the corresponding ROC curves in Fig.~\ref{Fig_5}. From the results, we can find input data with scale $2, 3, 4$ improve the anti-spoofing consistently over different frames. These results further show that the background region is useful for distinguishing genuine and fake face images. However, the improvement may discount when containing too much background.

\subsubsection{Test on REPLAY-ATTACK dataset}

\begin{figure*}
\begin{minipage}[b]{0.3\linewidth}
\centering
\centerline{\epsfig{figure=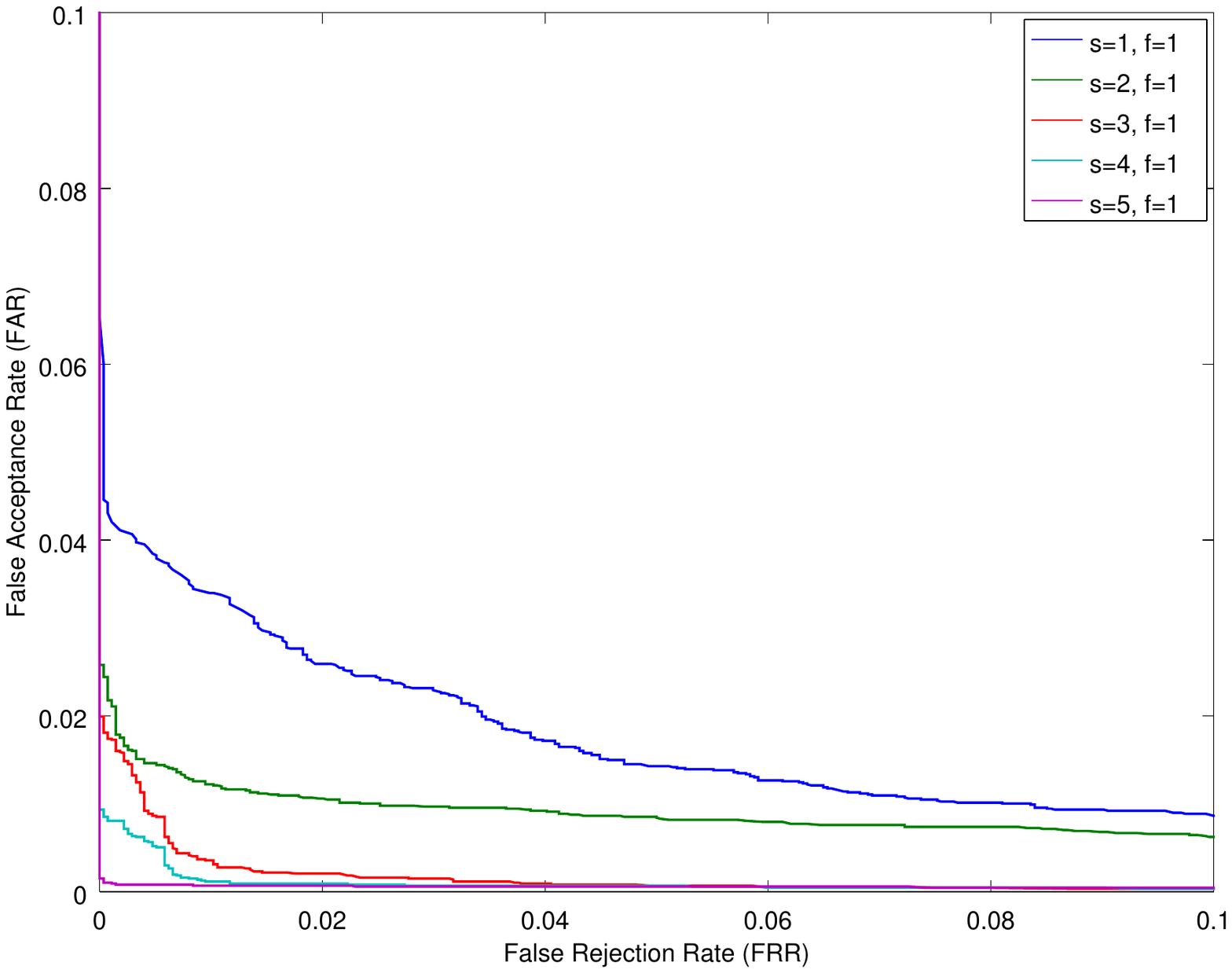,height=3.9cm, trim = 0 3in 0 3in 0}}
\end{minipage}
\hfill
\begin{minipage}[b]{0.3\linewidth}
\centering
\centerline{\epsfig{figure=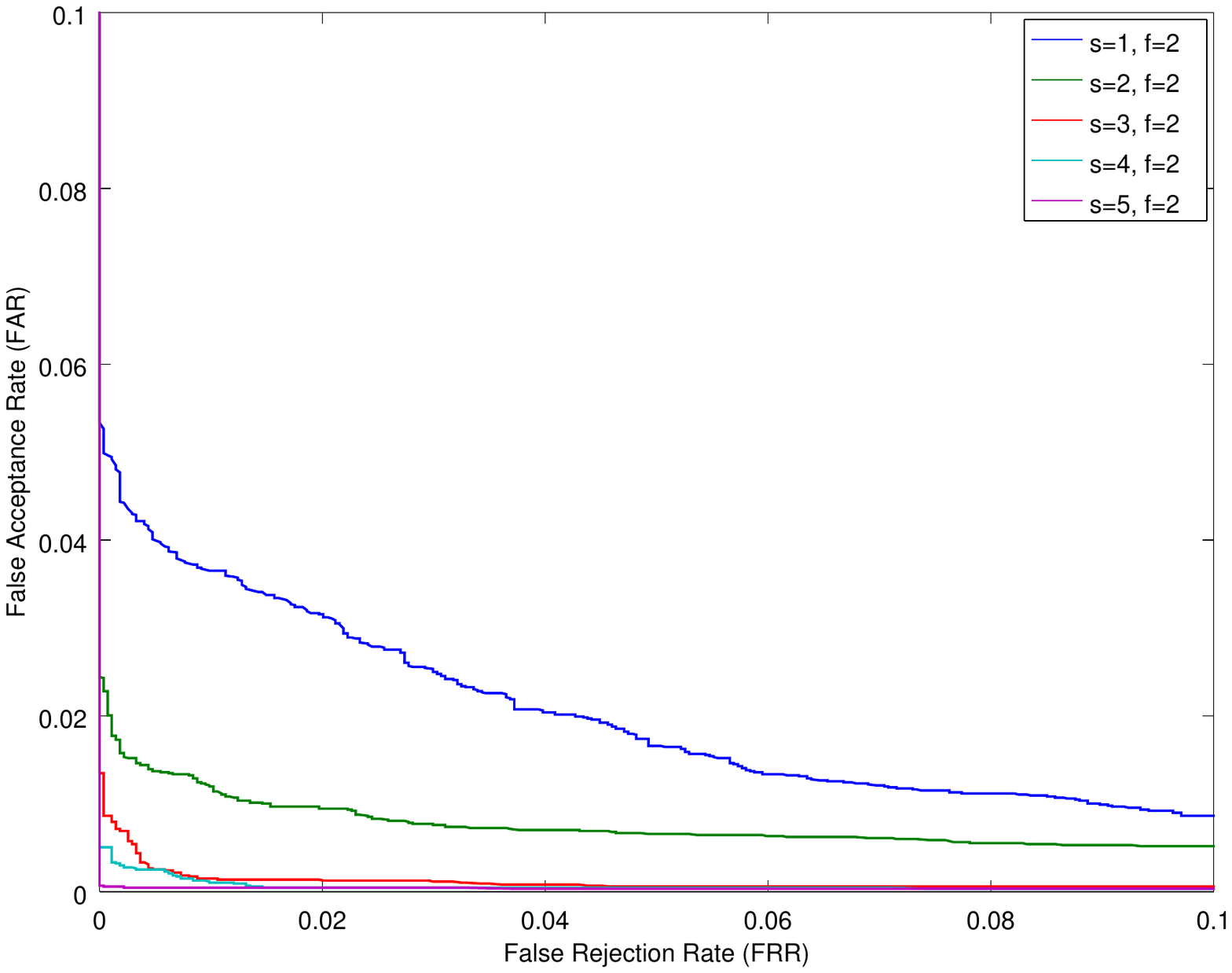,height=3.9cm, trim = 0 3in 0 3in 0}}
\end{minipage}
\hfill
\begin{minipage}[b]{0.3\linewidth}
\centering
\centerline{\epsfig{figure=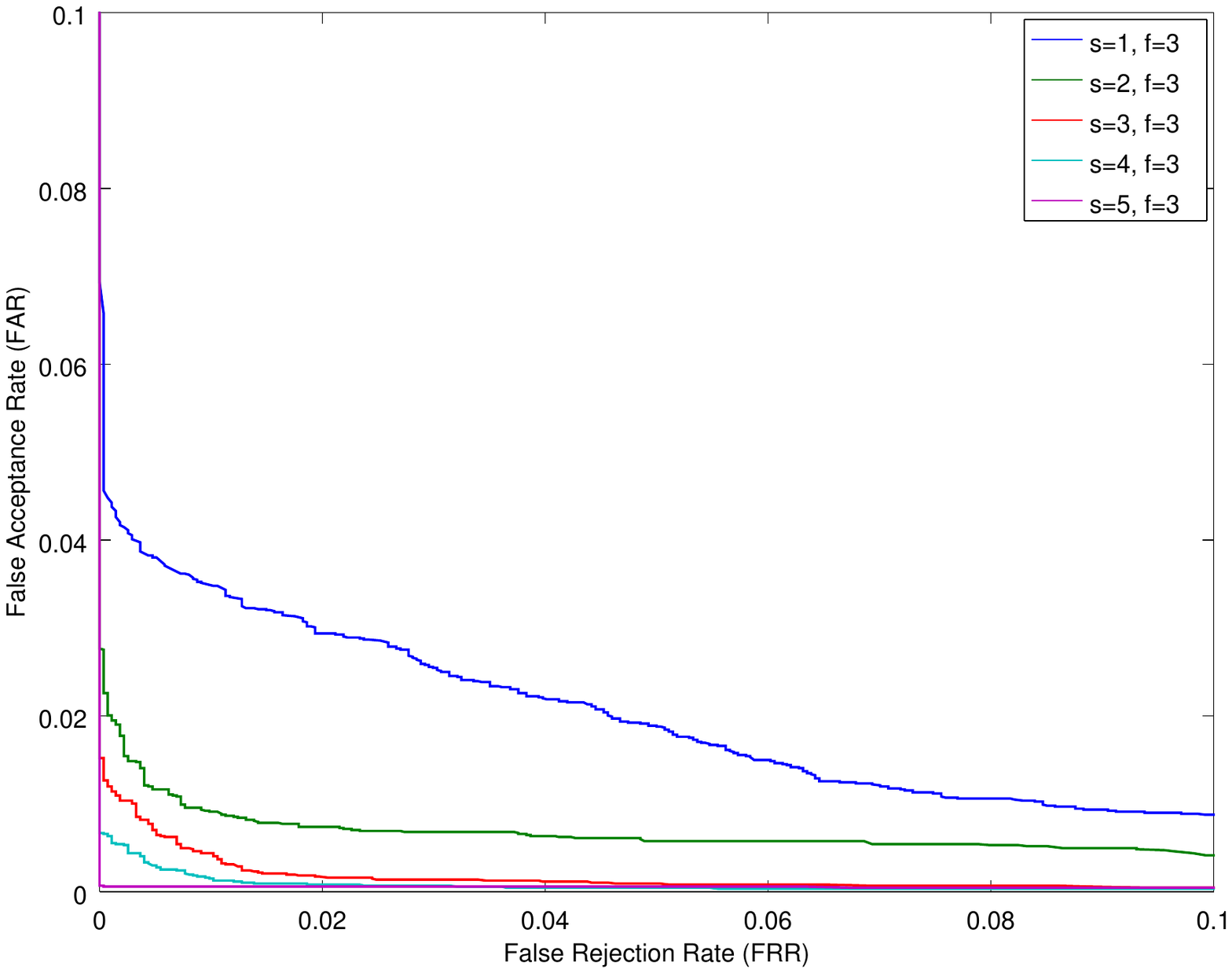,height=3.9cm, trim = 0 3in 0 3in 0}}
\end{minipage}
\caption{ROC curves for different data augmentations on REPLAY-ATTACK dataset. The display order is similar to Fig.~\ref{Fig_5}.}
\label{Fig_6}
\end{figure*}

On Replay-Attack dataset, we also report the performance under 15 scenarios. In Table~\ref{TB_REPLAY_ATTACK_INTRA_OURS}, we present the EERs and HTERs for different image scales and frame numbers used. In the table, we find some differences from CASIA dataset: (1) the lowest HTER occurs at scale $=1$, instead of a larger scale; (2) EERs on development set are larger than HTER on test set. These differences are mainly caused by the bias between development and test sets. On the test set, the models achieve accuracies all above 97\%, which are 2\%-5\% higher than the development set. To evaluate the performance more comprehensively, we draw the ROC curves for all scenarios in Fig.~\ref{Fig_6}. Accordingly, with the increase of scale, the performance improve gradually. When trained using input data with scale 5, the anti-spoofing achieves nearly perfect performance.

\subsubsection{Comparison}

For comparison, Table~\ref{TB_CASIA_INTRA_PREVIOUS} shows the intra-test results on CASIA dataset in \cite{Face_Anti_Spoofing_Pereira_2013}. As mentioned before, we use the same protocols as \cite{Face_Anti_Spoofing_Pereira_2013} for fair comparison. In \cite{Face_Anti_Spoofing_Pereira_2013}, the lowest EER is 21.59 on CASIA dataset, while it is 4.64 in our paper. Meanwhile, the lowest HTER in \cite{Face_Anti_Spoofing_Pereira_2013} is more than 4 times of ours. Such drastic improvements also occur on the REPLAY-ATTACK dataset. Such promising results indicate that the CNN can learn more discriminative features from input data compared with various hand-crafted features.

\begin{table}[!htb]
\caption{Intra-test Results on CASIA and Replay-Attack dataset in \cite{Face_Anti_Spoofing_Pereira_2013}.}
\label{TB_CASIA_INTRA_PREVIOUS}
\centering
\begin{tabular}{|c|c|c|c|c|c|c|}
\cline{2-7}
 \multicolumn{1}{c|}{}      & \multicolumn{6}{c|}{Feature}     \\
       \cline{2-7}
  \multicolumn{1}{c|}{}     & \multicolumn{2}{c|}{Correlation} & \multicolumn{2}{c|}{LBP-TOP$^{u2}_{8,8,8,1,1,1}$} & \multicolumn{2}{c|}{LBP$^{u2}_{8, 1}$} \\
\hline
Dataset       & dev & test & dev & test & dev & test \\
       \hline
CASIA   & 26.65 & 30.33 & 21.59 & 23.75 & 24.63 & 23.19 \\
\hline
Replay-Attack  & 11.66 & 11.79 & 8.17 & 8.51 & 14.41 & 15.45 \\
\hline
\end{tabular}
\end{table}

\subsection{Results of Inter-test}

For face anti-spoofing, its adaptation ability from one dataset to another is important for practical applications. In this part, we will evaluate such ability of CNN models. Similar to \cite{Face_Anti_Spoofing_Pereira_2013}, we first train our model using the training set from CASIA dataset, and then validate and test it using Replay-Attack dataset. This procedure is then inverted, i.e., using Replay-Attack dataset for training and CASIA dataset for development and testing. In this inter-test procedure, training data is used to tune the CNN models and train SVM classifiers, which are then used to extract features and perform evaluation on the development and test sets, respectively. When extracting features from development and test sets, the mean of training data is used for centralizations of development and testing data.

\begin{figure*}
\begin{minipage}[b]{0.3\linewidth}
\centering
\centerline{\epsfig{figure=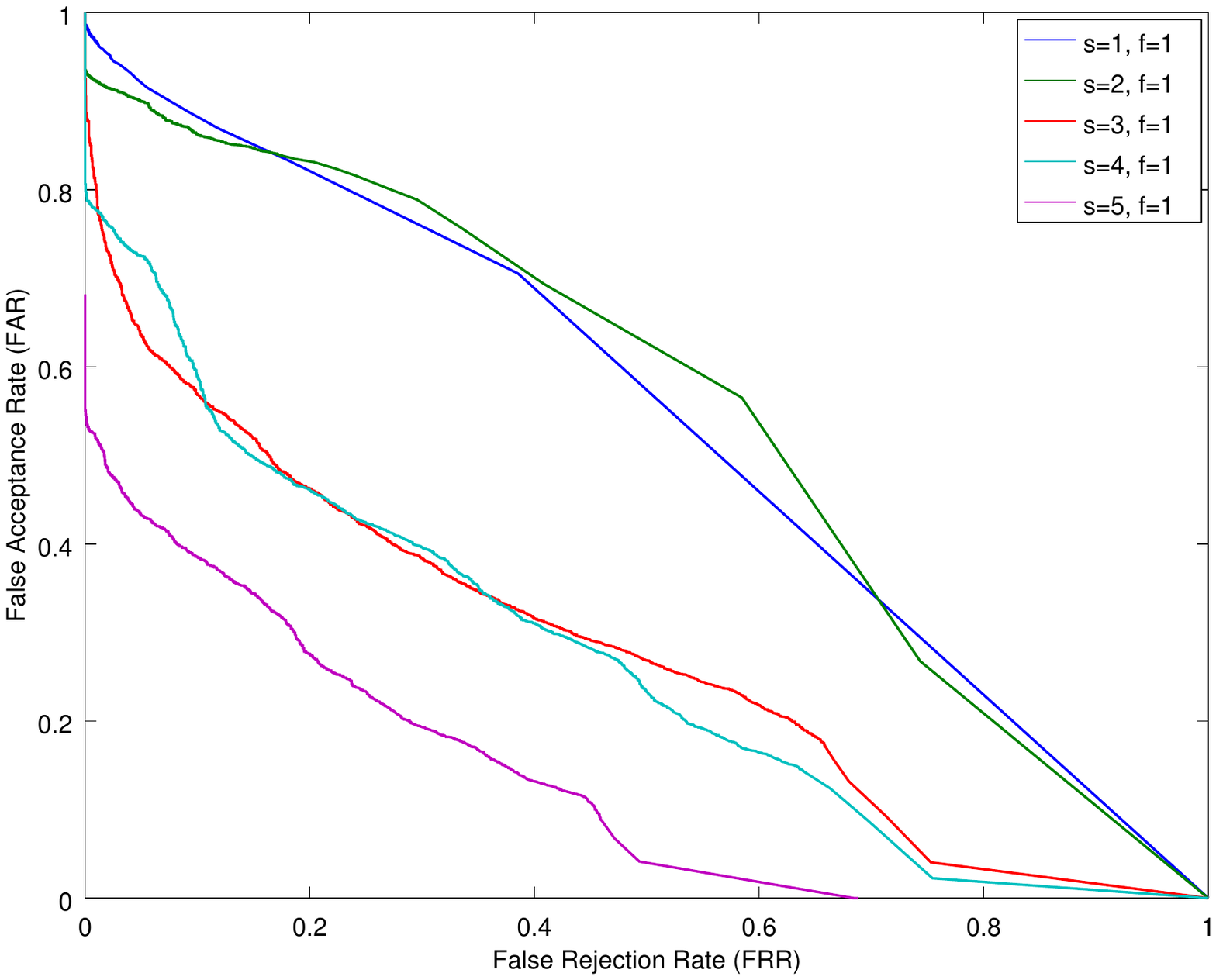,height=3.9cm, trim = 0 3in 0 3in 0}}
\end{minipage}
\hfill
\begin{minipage}[b]{0.3\linewidth}
\centering
\centerline{\epsfig{figure=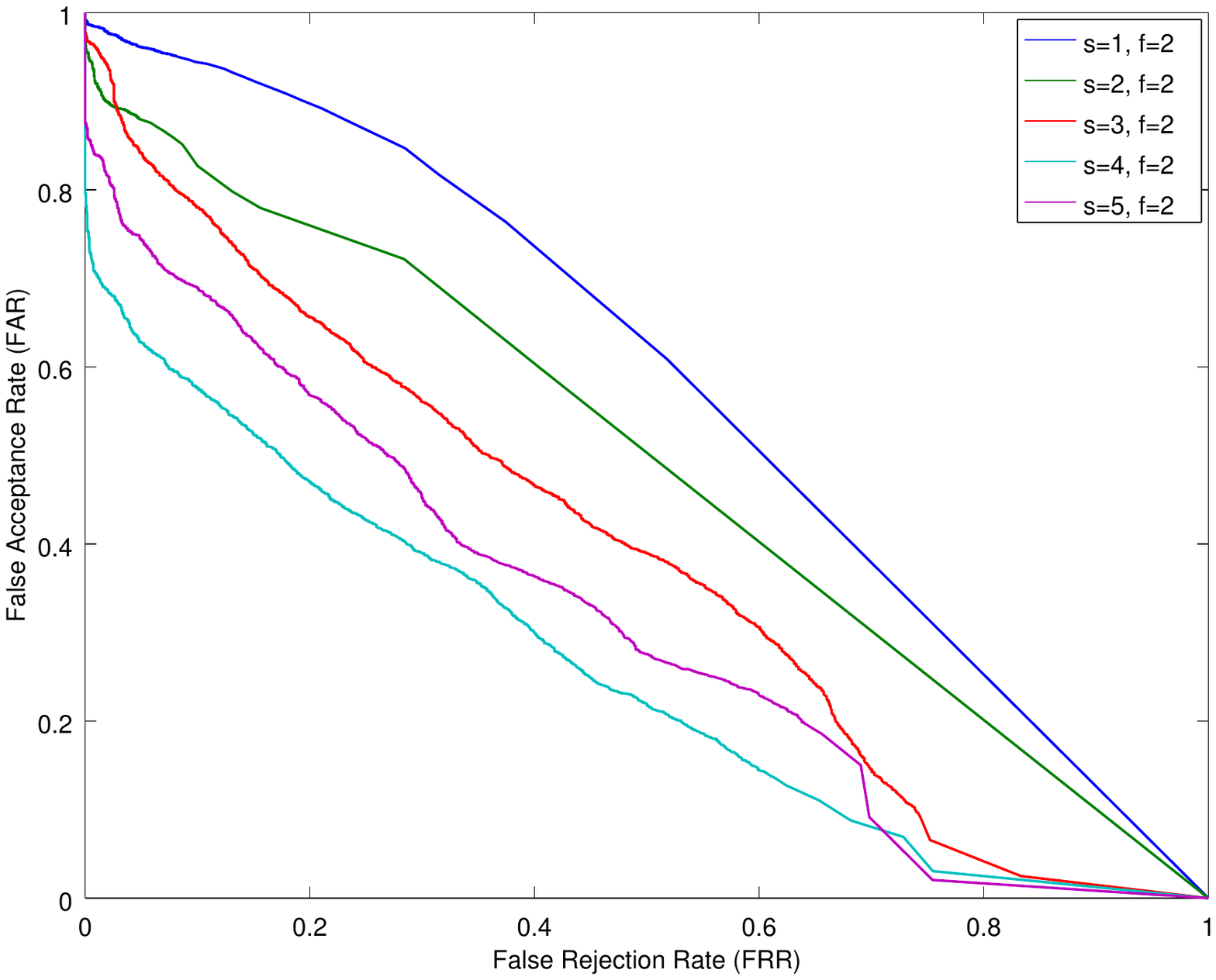,height=3.9cm, trim = 0 3in 0 3in 0}}
\end{minipage}
\hfill
\begin{minipage}[b]{0.3\linewidth}
\centering
\centerline{\epsfig{figure=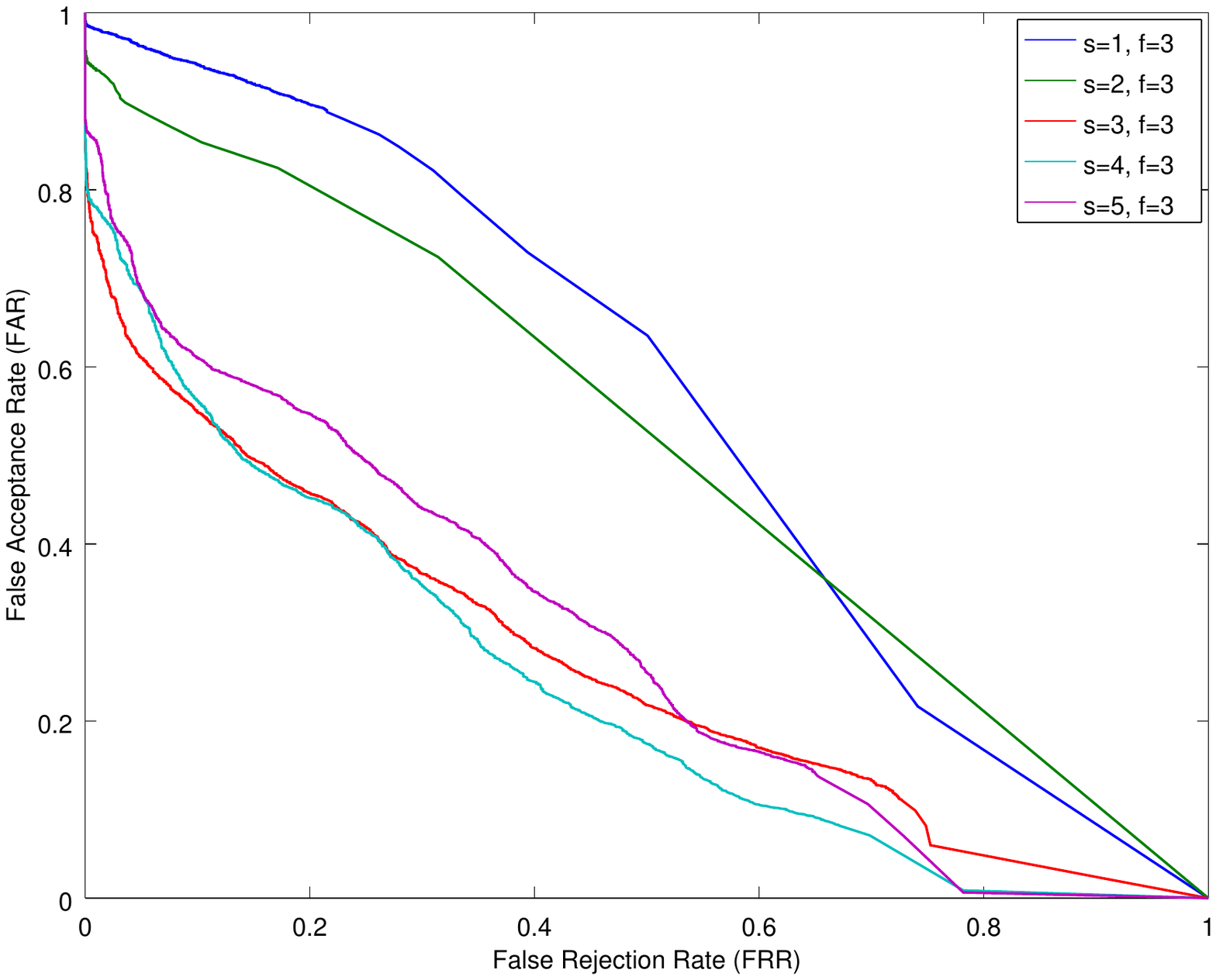,height=3.9cm, trim = 0 3in 0 3in 0}}
\end{minipage}
\caption{ROC curves of inter-test for different inputs. The models are trained on CASIA dataset, and developed and tested on REPLAY-ATTACK dataset. The display order is similar to Fig.~\ref{Fig_5}.}
\label{Fig_7}
\end{figure*}

\begin{figure*}
\begin{minipage}[b]{0.3\linewidth}
\centering
\centerline{\epsfig{figure=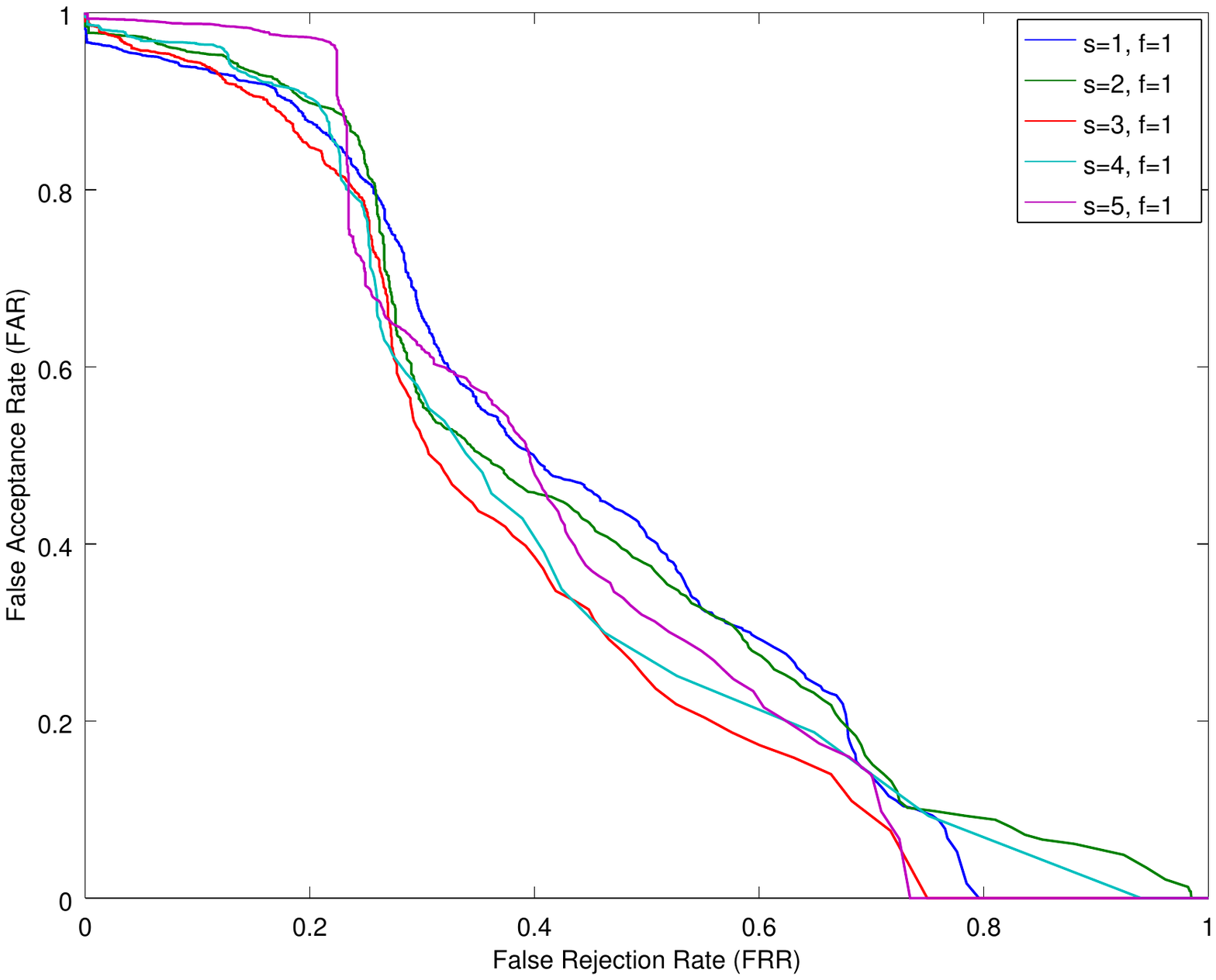,height=3.9cm, trim = 0 3in 0 3in 0}}
\end{minipage}
\hfill
\begin{minipage}[b]{0.3\linewidth}
\centering
\centerline{\epsfig{figure=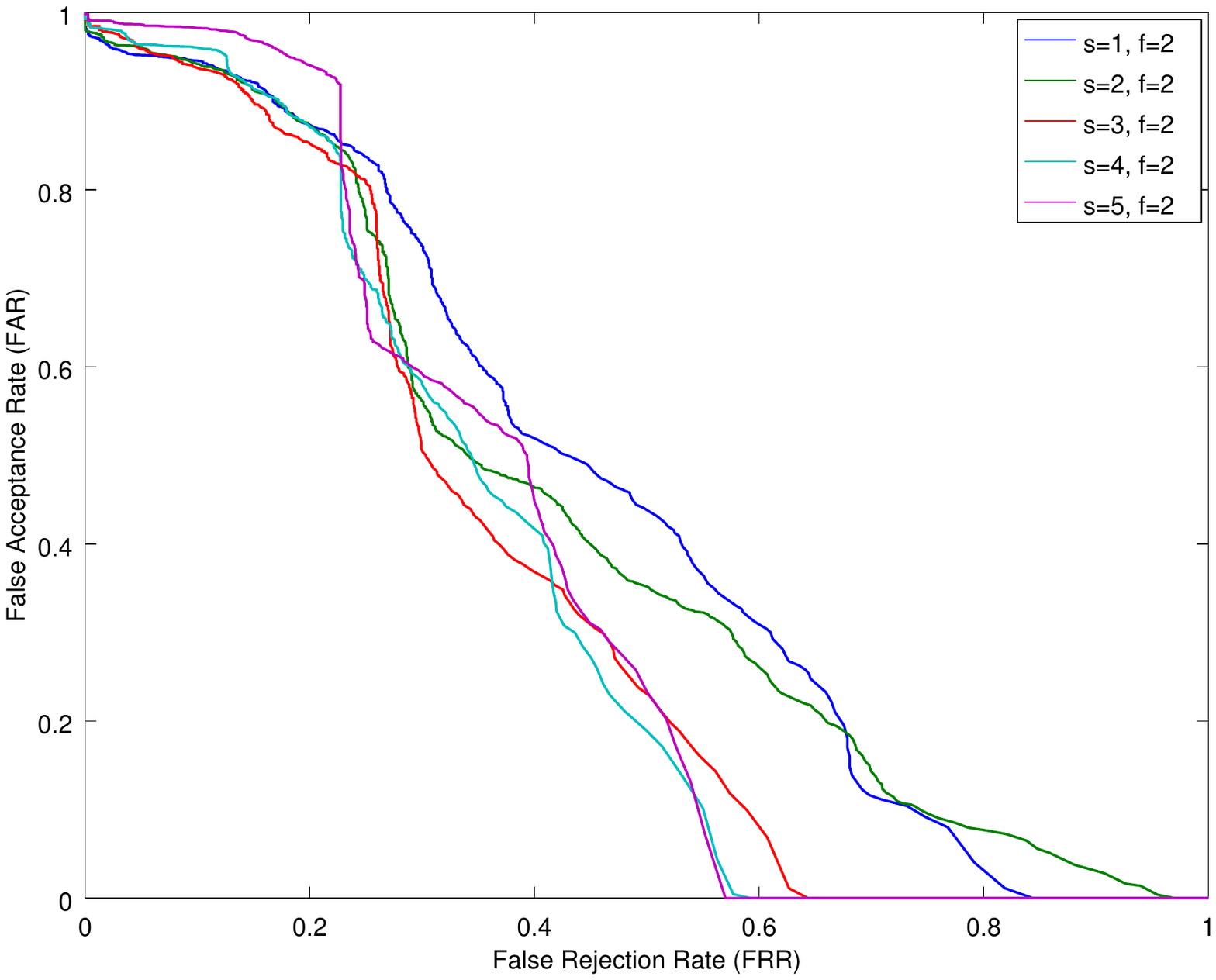,height=3.9cm, trim = 0 3in 0 3in 0}}
\end{minipage}
\hfill
\begin{minipage}[b]{0.3\linewidth}
\centering
\centerline{\epsfig{figure=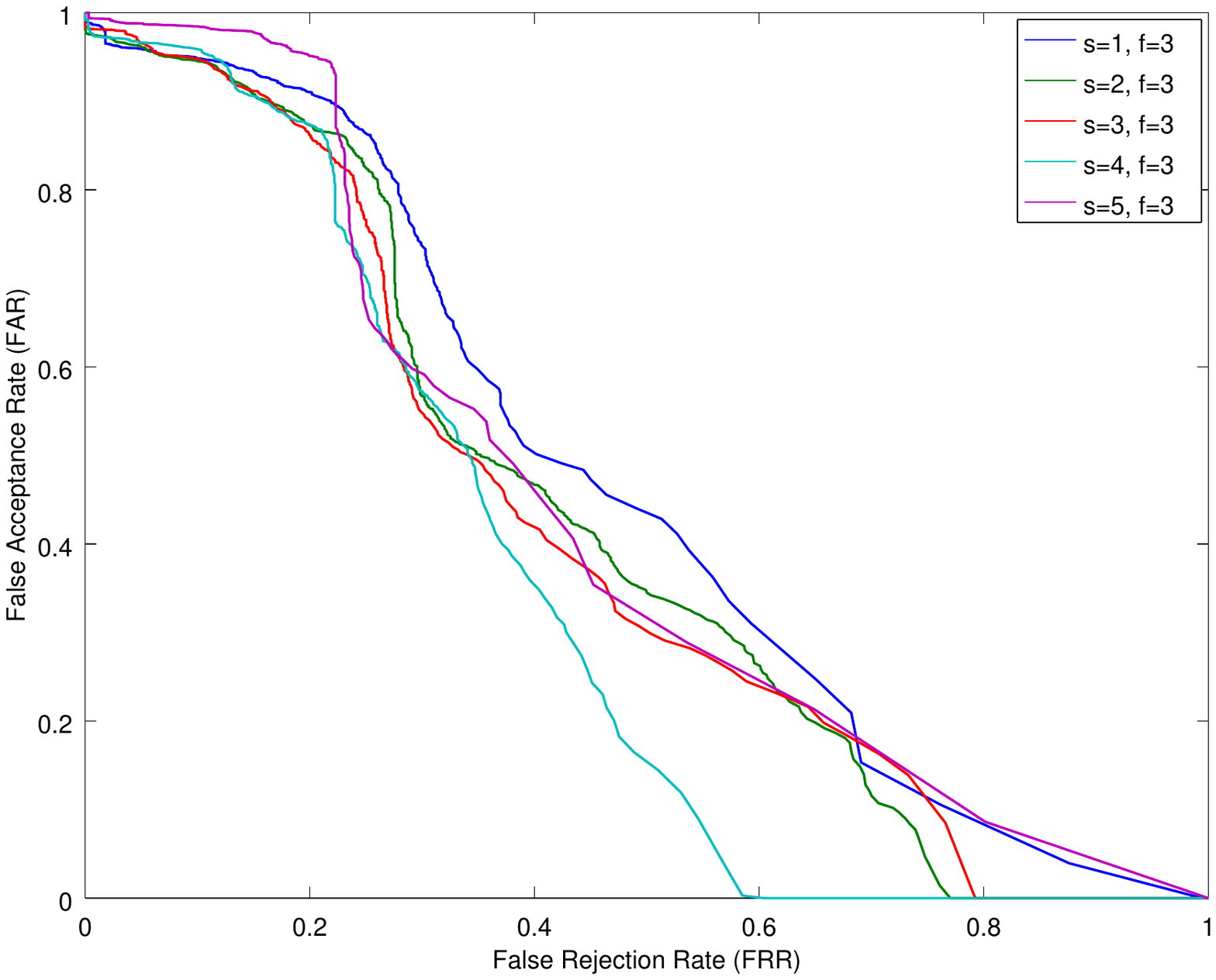,height=3.9cm, trim = 0 3in 0 3in 0}}
\end{minipage}
\caption{ROC curves of inter-test. The models are trained on REPLAY-ATTACK dataset, and developed and tested on CASIA dataset. The display order is similar to Fig.~\ref{Fig_7}.}
\label{Fig_8}
\end{figure*}

As shown in Table~\ref{TB_INTER_OURS}, the top four lines show the EERs and HTERs when using CASIA for training and REPLAY-ATTACK for testing; and the bottom four lines present results from the inverse. For comparison, we show the performance of \cite{} in Table~\ref{TB_INTER_PREVIOUS}. Accordingly, the performance of our method is analogous to that in \cite{Face_Anti_Spoofing_Pereira_2013} when the scale is 1, which indicates that both hand-crafted and learned features are incapable of capturing common cues from face regions across datasets. However, with the scale increased, such a situation changes. For the REPLAY-ATTACK dataset, the testing performance improves gradually, and the lowest HTER approaches to 23.78 when using one frame with scale 5. Similarly on the CASIA dataset, the HTER also decreases when input data contains backgrounds. The lowest HTER is 38.11 when using three frames with scale 4 as input. In Fig.~\ref{Fig_7} and \ref{Fig_8}, we show the ROC curves for different inter-test scenarios.
  
According the experiments in this part, we can find the cross-datasets anti-spoofing is far from satisfactory. Due to different devices, illuminations, races, etc., there are some inevitable biases among two datasets. In this case, the inter-dataset can hardly obtain analogous performance to the intra-test situation. Fortunately, we find in our experiments that background regions can boost the generalization ability of anti-spoofing model.

\begin{table*}[!htb]
\caption{Inter-test results on CASIA and REPLAY-ATTACK datasets.}
\label{TB_INTER_OURS}
\centering
\begin{tabular}{|c|c|c|c|c|c|c|c|c|c|c|c||c|c|}
\cline{3-14}
\multicolumn{2}{c|}{}       & \multicolumn{12}{c|}{Scale}     \\
\cline{3-14}
\multicolumn{2}{c|}{} & \multicolumn{2}{c|}{1} & \multicolumn{2}{c|}{2} & \multicolumn{2}{c|}{3} & \multicolumn{2}{c|}{4} & \multicolumn{2}{c|}{5} &  \multicolumn{2}{c|}{Mean}\\
\cline{3-14}
\multicolumn{2}{c|}{} & dev & test & dev & test & dev & test & dev & test & dev & test & dev & test \\
\hline
\multirow{3}{*}{Frame} & 1 & 48.22 & 48.80 & 50.17 & 50.56 & 41.39 & 38.02 & 33.16 & 36.51 & 38.80 & 23.78 & 42.25 & 39.53 \\ 

                       & 2 & 49.60 & 51.58 & 45.85 & 46.39 & 49.20 & 44.02 & 35.53 & 34.93 & 42.82 & 38.21 & 44.60 & 43.03 \\ 

                       & 3 & 51.37 & 55.16 & 48.17 & 47.34 & 39.25 & 34.90 & 30.63 & 32.14 & 33.47 & 38.03 & 40.58 & 41.51 \\ 
\hline
\hline
& Mean & 49.73 & 51.91 & 48.06 & 48.10 & 43.28 & 38.98 & 33.10 & 34.53 & 38.36 & 33.34 & 42.48 & 41.36 \\
\hline
\end{tabular}

\begin{tabular}{|c|c|c|c|c|c|c|c|c|c|c|c||c|c|}
\hline
\hline
\multirow{3}{*}{Frame} & 1 & 45.76 & 45.49 & 43.02 & 43.63 & 39.33 & 39.31 & 39.97 & 40.03 & 41.90 & 42.42 & 42.00 & 42.18 \\ 

                       & 2 & 46.48 & 46.81 & 42.70 & 42.88 & 38.29 & 38.38 & 40.24 & 40.49 & 39.88 & 40.88 & 41.52 & 41.89 \\ 

                       & 3 & 46.07 & 46.43 & 43.25 & 42.97 & 40.88 & 40.78 & 38.91 & 38.11 & 41.18 & 41.91 & 42.06 & 42.04 \\ 
\hline
\hline
& Mean & 46.10 & 46.24 & 42.99 & 43.16 & 39.50 & 39.49 & 39.71 & 39.54 & 40.10 & 41.74 & 41.86 & 42.04 \\
\hline
\end{tabular}
\end{table*}

\begin{table}[!htb]
\caption{Inter-test results on CASIA and Replay-Attack datasets in \cite{Face_Anti_Spoofing_Pereira_2013}.}
\label{TB_INTER_PREVIOUS}
\centering
\begin{tabular}{|c|c|c|c|c|c|c|}
\cline{2-7}
\multicolumn{1}{c|}{}      & \multicolumn{6}{c|}{Feature}     \\
\cline{2-7}
\multicolumn{1}{c|}{}      & \multicolumn{2}{c|}{Correlation} & \multicolumn{2}{c|}{LBP-TOP$^{u2}_{8,8,8,1,1,1}$} & \multicolumn{2}{c|}{LBP$^{u2}_{8, 1}$} \\
       \hline
Training Set       & dev & test & dev & test & dev & test \\
       \hline
CASIA   & 50.23 & 50.25 & 48.97 & 50.64 & 44.97 & 47.05 \\
\hline
Replay-Attack  & 47.72 & 48.28 & 60.00 & 61.33 & 57.32 & 57.90 \\
\hline
\end{tabular}
\end{table}

\subsection{Results on Combined Datasets}

\begin{figure*}
\begin{minipage}[b]{0.3\linewidth}
\centering
\centerline{\epsfig{figure=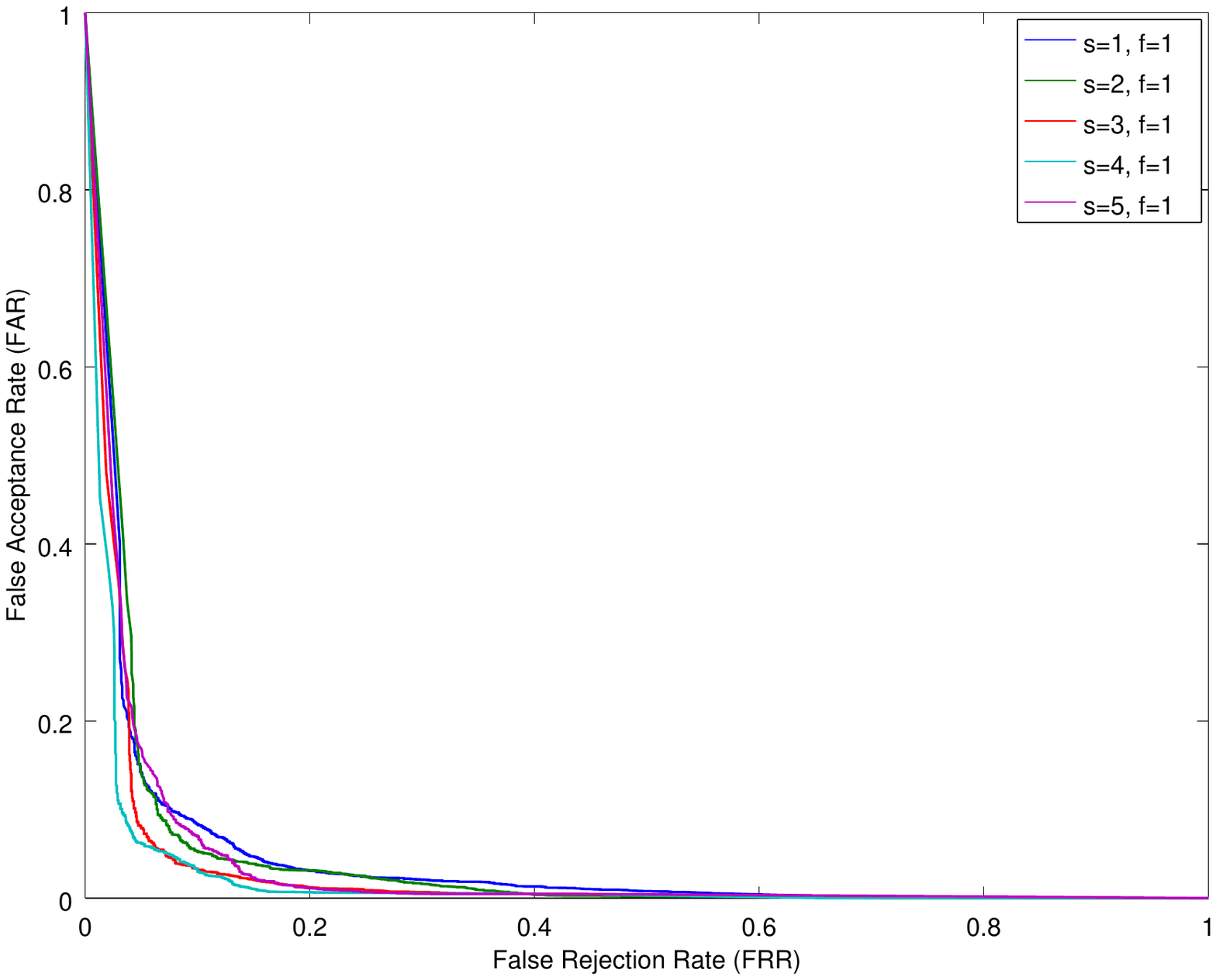,height=3.9cm, trim = 0 3in 0 3in 0}}
\end{minipage}
\hfill
\begin{minipage}[b]{0.3\linewidth}
\centering
\centerline{\epsfig{figure=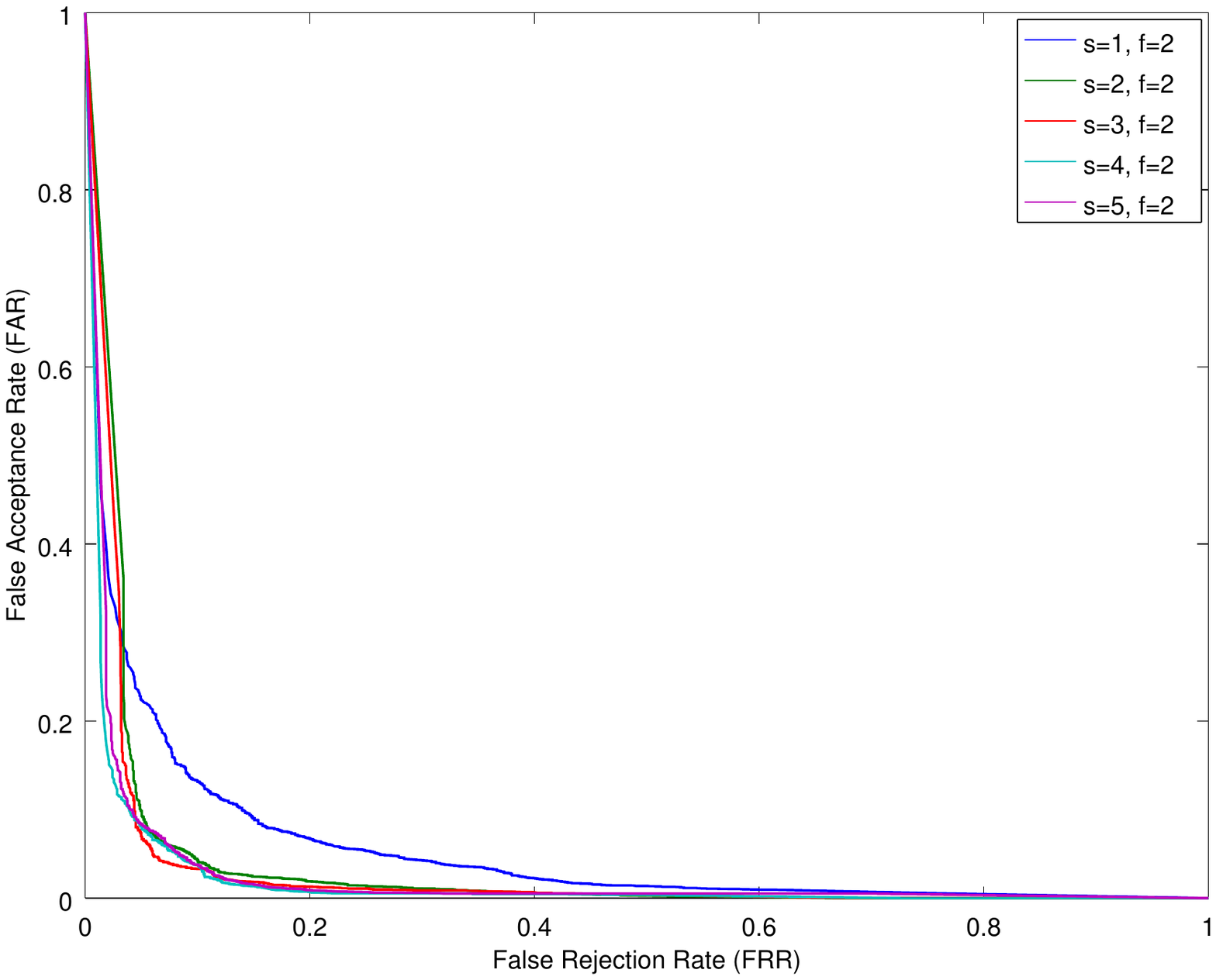,height=3.9cm, trim = 0 3in 0 3in 0}}
\end{minipage}
\hfill
\begin{minipage}[b]{0.3\linewidth}
\centering
\centerline{\epsfig{figure=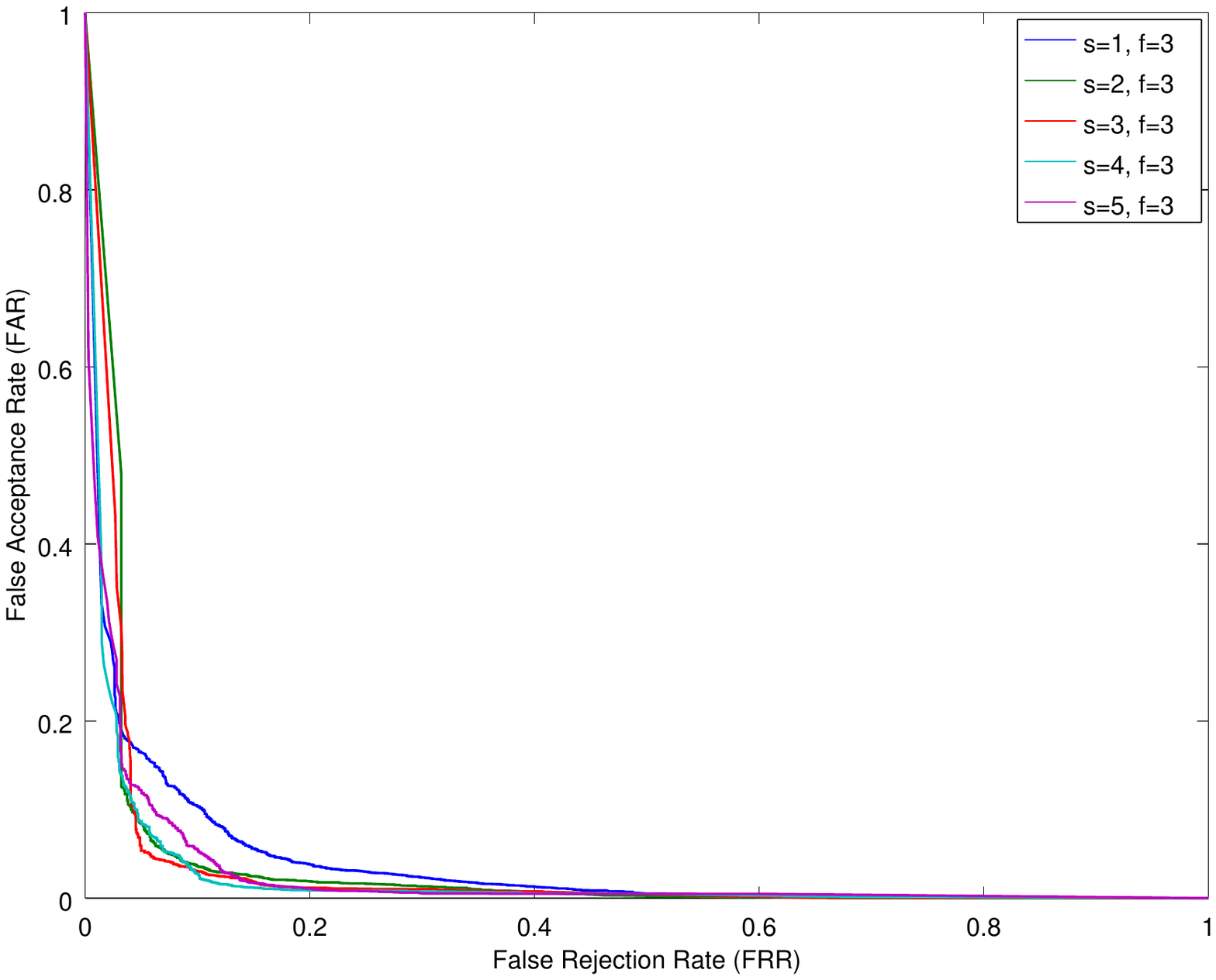,height=3.9cm, trim = 0 3in 0 3in 0}}
\end{minipage}
\\
\\
\begin{minipage}[b]{0.3\linewidth}
\centering
\centerline{\epsfig{figure=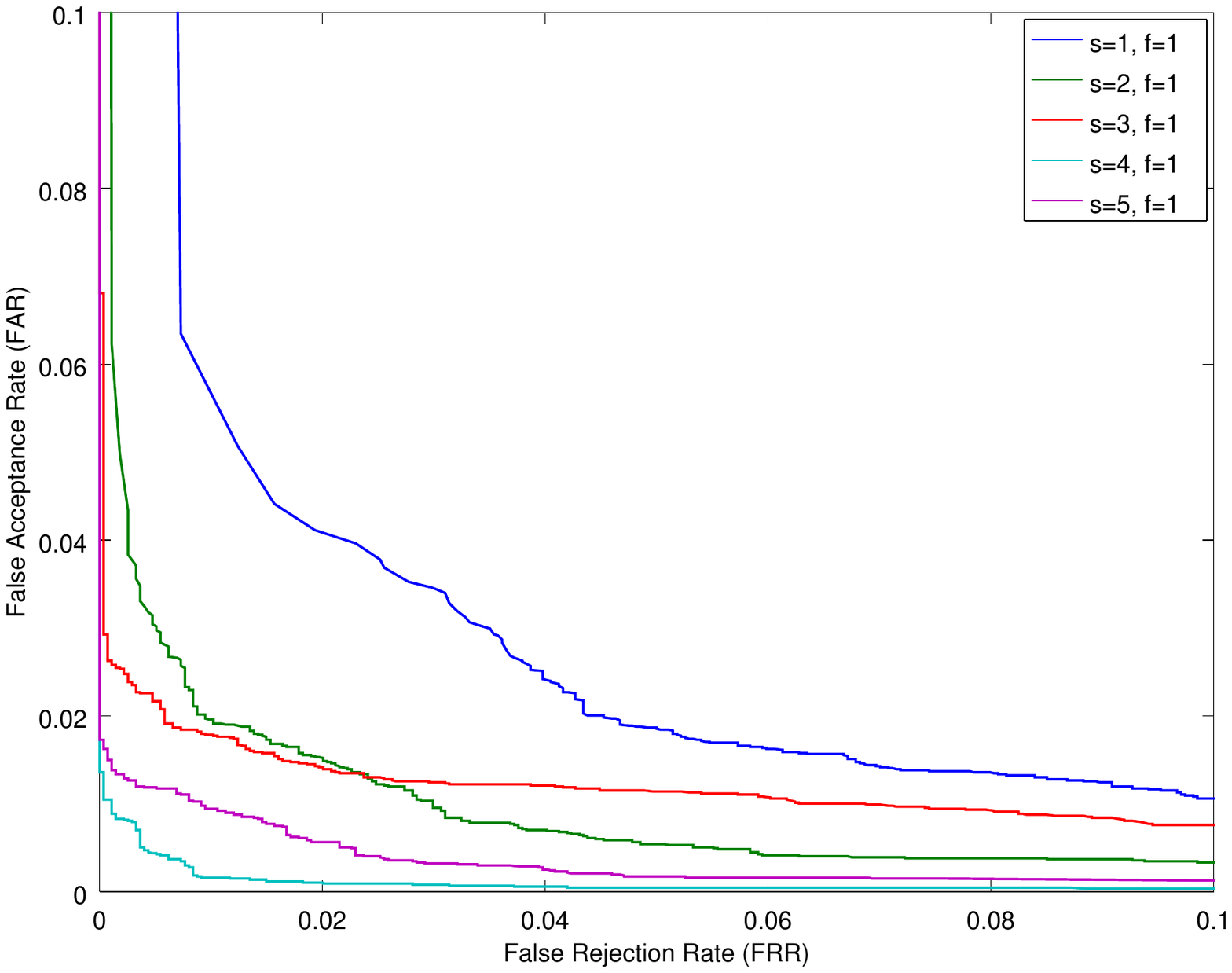,height=3.9cm, trim = 0 3in 0 3in 0}}
\end{minipage}
\hfill
\begin{minipage}[b]{0.3\linewidth}
\centering
\centerline{\epsfig{figure=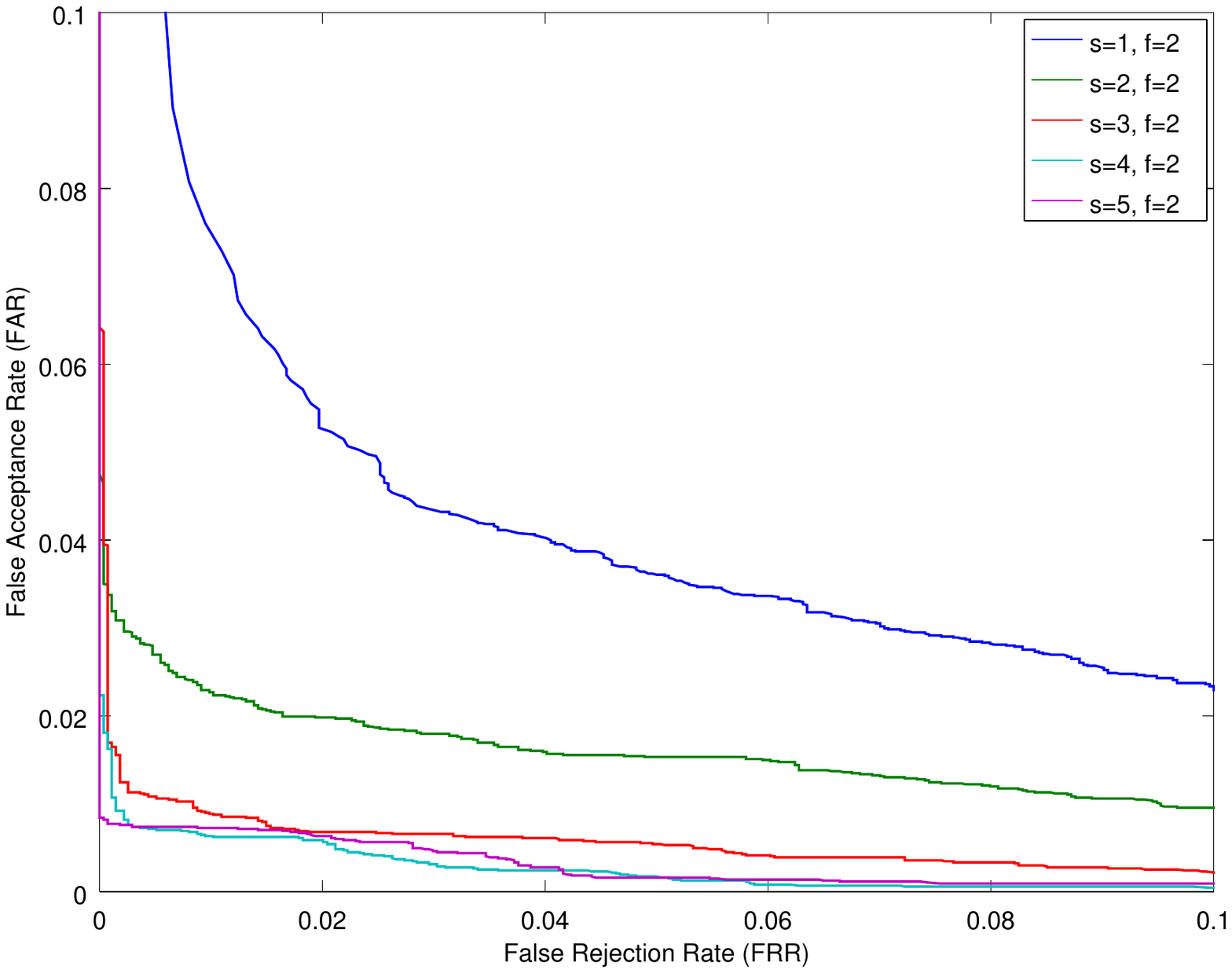,height=3.9cm, trim = 0 3in 0 3in 0}}
\end{minipage}
\hfill
\begin{minipage}[b]{0.3\linewidth}
\centering
\centerline{\epsfig{figure=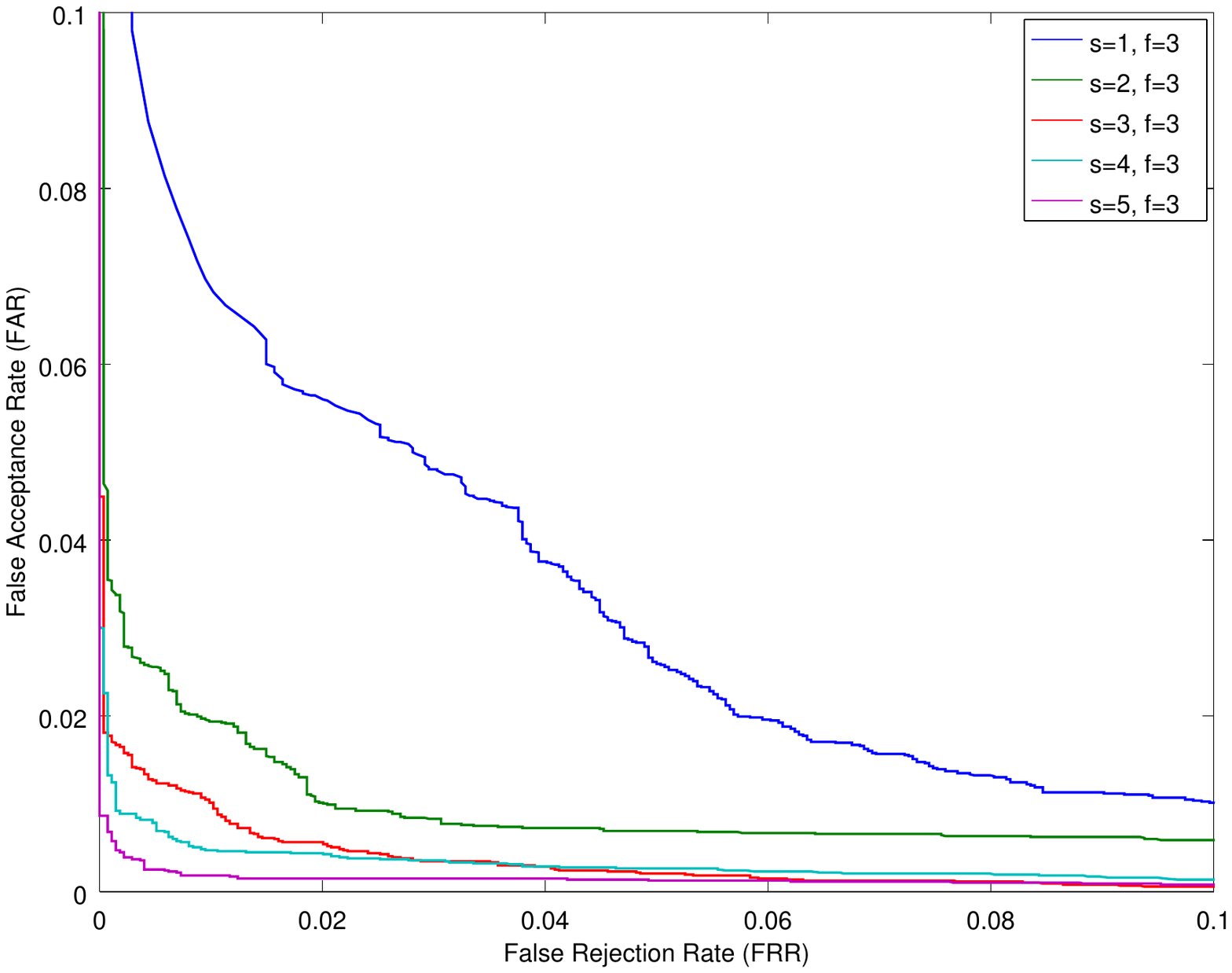,height=3.9cm, trim = 0 3in 0 3in 0}}
\end{minipage}
\caption{ROC curves for data combination protocol. The top three figures show models tested on CASIA dataset, and the bottom three figures show performance of models tested on REPLAY-ATTACK dataset.}
\label{Fig_9}
\end{figure*}

In this part, we assume training data from two datasets are available. As the protocols proposed in \cite{Face_Anti_Spoofing_Pereira_2013}, our models are trained and developed on combined datasets, and then evaluated on each dataset. In Table~\ref{TB_COMBINED_OURS}, we show the EERs and HTERs of all scenarios. Compared with the results in \cite{Face_Anti_Spoofing_Pereira_2013} (Table~\ref{TB_COMBINED_PREVIOUS}), our method achieves much better result on both datasets. On the CASIA dataset, we obtain comparable performance to the intra-test; On the REPLAY-ATTACK dataset, the average HTERs are less than 1\% when scale $=$ 4 and 5. In Fig.~\ref{Fig_9}, we show ROC curves of different cases. We can find the models trained using samples from two dataset perform similarly to those in the intra-test, which illustrates that CNN are able to learn common features from such two datasets. Moreover, compared with \cite{Face_Anti_Spoofing_Pereira_2013}, the performance is not biased as much between two datasets due to the powerful feature learning ability of CNN.

\subsection{Discussion}

Thus far, we have evaluated our method in various cases. From the experimental results, we show that the proposed method has achieved much better performance in all testing protocols. These encouraging results prove the power of CNN once again, but the first time in face anti-spoofing. Compared with previous hand-crafted features, such a data-driven feature designing rules out the empirical part, and come up with some more efficient features for face anti-spoofing. Moreover, it should be pointed out that we did not pay attentions on parameter selecting for CNN, but we believe that a better model can be obtained after some efforts on it. 

Beyond the powerful feature learning toolkit, we also proposed many data augmentation strategies, including spatial and temporal augmentations. By augmenting the input data, we further improve face anti-spoofing performance in all protocols. These improvements suggests that background region is indeed helpful for face anti-spoofing to some extent when using CNN-learned or hand-crafted features \cite{Face_Anti_Spoofing_JianweiYang_2013}. Though the improvements are seen on both datasets, there are some difference to be pointed out. Specifically, on the CASIA dataset, the best scale is 3, while 5 for REPLAY-ATTACK dataset. This inconsistency can be explained by the input data partially. In the CASIA dataset, all sequences of spoofing attacks contain real-world backgrounds as real-access sequences. However, all background regions are filled by fake photos in REPLAY-ATTACK dataset. As a result, when the scale is too large, genuine and fake samples in CASIA dataset become more similar rather than different, whereas they are more discriminative on REPLAY-ATTACK dataset. At this point, we argue that face anti-spoofing should not be regarded as a classification problem on faces, but one on the regions where fake faces are shown.

\begin{table*}[!htb]
\caption{Results from data combination protocol. The models are trained on two datasets, and then developed and tested on each dataset separately.}
\label{TB_COMBINED_OURS}
\centering
\begin{tabular}{|c|c|c|c|c|c|c|c|c|c|c|c||c|c|}
\cline{3-14}
\multicolumn{2}{c|}{}       & \multicolumn{12}{c|}{Scale}     \\
\cline{3-14}
\multicolumn{2}{c|}{} & \multicolumn{2}{c|}{1} & \multicolumn{2}{c|}{2} & \multicolumn{2}{c|}{3} & \multicolumn{2}{c|}{4} & \multicolumn{2}{c|}{5} &  \multicolumn{2}{c|}{Mean}\\
\cline{3-14}
\multicolumn{2}{c|}{} & dev & test & dev & test & dev & test & dev & test & dev & test & dev & test \\
\hline
\multirow{6}{*}{Frame} & \multirow{2}{*}{1} & \multirow{2}{*}{10.89} & 13.29 & \multirow{2}{*}{4.85} & 9.04 & \multirow{2}{*}{1.77} & 6.08 & \multirow{2}{*}{1.29} & 6.30 & \multirow{2}{*}{2.65} & 8.62 & \multirow{2}{*}{4.29} & 8.67 \\ 
                      &  &  & 3.25 & & 1.57 & & 1.34 & & 0.70 & & 1.19 & & 1.61 \\ 
\cline{2-14}                   
                      & \multirow{2}{*}{2} & \multirow{2}{*}{11.28} & 13.71 & \multirow{2}{*}{3.56} & 6.58 & \multirow{2}{*}{2.32} & 5.81 & \multirow{2}{*}{1.55} & 6.59 & \multirow{2}{*}{2.79} & 6.74 & \multirow{2}{*}{4.30} & 7.89 \\
                      &  &  & 3.58 & & 1.61 & & 1.36 & & 0.60 & & 0.95 & & 1.62 \\
\cline{2-14}                  
                       & \multirow{2}{*}{3} & \multirow{2}{*}{10.12} & 11.21 & \multirow{2}{*}{3.85} & 8.27 & \multirow{2}{*}{2.05} & 5.37 & \multirow{2}{*}{2.13} & 6.54 & \multirow{2}{*}{2.07} & 7.72 & \multirow{2}{*}{4.04} & 7.82 \\ 
                       &  &  & 3.90 &  & 1.52 &  & 1.03 &  & 0.68 &  & 0.40 & & 1.51 \\ 
\hline
\hline
& \multirow{2}{*}{Mean} & \multirow{2}{*}{10.76} & 12.74 & \multirow{2}{*}{4.09} & 7.96 & \multirow{2}{*}{2.05} & 5.75 & \multirow{2}{*}{1.66} & 6.48 & \multirow{2}{*}{2.50} & 7.69 & \multirow{2}{*}{4.21} & 8.13 \\
&  &  & 3.58 & & 1.57 & & 1.24 & & 0.66 & & 0.85 & & 1.58 \\
\hline
\end{tabular}
\end{table*}

\begin{table}[!htb]
\caption{Results of data combination protocol on CASIA and Replay-Attack datasets in \cite{Face_Anti_Spoofing_Pereira_2013}.}
\label{TB_COMBINED_PREVIOUS}
\centering
\begin{tabular}{|c|c|c|c|c|c|c|}
\cline{2-7}
\multicolumn{1}{c|}{}      & \multicolumn{6}{c|}{Feature}     \\
       \cline{2-7}
 \multicolumn{1}{c|}{}     & \multicolumn{2}{c|}{Correlation} & \multicolumn{2}{c|}{LBP-TOP$^{u2}_{8,8,8,1,1,1}$} & \multicolumn{2}{c|}{LBP$^{u2}_{8, 1}$} \\
       \hline
Testing Set      & dev & test & dev & test & dev & test \\ 
       \hline
CASIA   & \multirow{2}{*}{12.18} & 43.30 & \multirow{2}{*}{14.29} & 42.04 & \multirow{2}{*}{20.45} & 45.92 \\
Replay-Attack  & & 24.14 & & 10.67 & & 10.07 \\
\hline
\end{tabular}
\end{table}

\section{CONCLUSIONS AND FUTURE WORKS}

In this paper, we have proposed to use CNN to learn features for face anti-spoofing. Upon the CNN model, we tried different data augmentation strategies. According to the experimental results. The proposed method make a significant improvement compared with previous works in different protocols. In the intra-test and combination protocols, we have achieved HTERs lower than 5\% on two datasets. In the inter-test protocol, there are also remarkable improvements. However, we must point out that the proposed method is still not able to obtain satisfactory performance in the inter-test protocol. As we discussed before, due to different capturing conditions, the biases among different datasets are inevitable. Towards this point, one of our future work is to find out a way to adapt the learned model to new data based on transfer learning. Also, integrating other cues, such as motions and shapes is another direction.

\bibliographystyle{ieee}
\bibliography{references}
\end{document}